\pdfoutput=1

\documentclass[letterpaper, 10 pt, conference]{ieeeconf}

\IEEEoverridecommandlockouts
\usepackage{multirow}
\usepackage{multicol}
\usepackage[table,xcdraw]{xcolor}
\usepackage{graphicx}
\usepackage[breaklinks,colorlinks,citecolor=flodarkpurple]{hyperref}
\usepackage[hypcap=false]{caption}
\usepackage{array}
\usepackage{colortbl}
\usepackage{wrapfig}
\usepackage{algorithm}
\usepackage{algpseudocode}
\usepackage{amsmath}
\usepackage{booktabs}
\usepackage{subcaption}
\usepackage{placeins}


\definecolor{lightgreenrow}{RGB}{226,239,218}
\definecolor{lightgrayrow}{RGB}{242,242,242}
\definecolor{headerpeach}{RGB}{253,233,217}

\newif\ifanonymous
\anonymousfalse  %

\newif\ifappendix
\appendixtrue

\usepackage{adjustbox}
\usepackage{amsmath}
\usepackage{amssymb}
\usepackage{array}  %
\usepackage{authblk}
\usepackage{booktabs}
\usepackage{capt-of}
\makeatletter
\let\NAT@parse\undefined
\makeatother
\usepackage{cite}
\usepackage{colortbl}
\usepackage{epsfig}
\usepackage{float}
\usepackage{inconsolata}  %
\usepackage{graphicx}
\usepackage{listings}
\usepackage{multicol}
\usepackage{multirow}
\usepackage{paralist}
\usepackage{pifont}
\usepackage{ragged2e} %
\usepackage{times}
\usepackage{tabularx}
\usepackage{xcolor}
\usepackage{svg}

\usepackage{subcaption}

\newcommand{\cooltitle}{\coolname\hspace{.1cm}}

\definecolor{flodarkpurple}{rgb}{0.288,0.1196,0.7}

\definecolor{amber}{rgb}{1.0, 0.75, 0.0}

\newcommand{\coolname}{\textit{iTeach}}

\ifanonymous
  \newcommand{\webpage}{[anonymized]}
\else
  \newcommand{\webpage}{https://irvlutd.github.io/iTeach}
\fi

\newcommand{\authorhref}[3][flodarkpurple]{\href{#2}{\color{#1}{#3}}}

\definecolor{navgold}{HTML}{ffaa00}

\title{\LARGE \bf
\coolname: In the Wild Interactive Teaching \\for Failure-Driven Adaptation of Robot Perception

}

\ifanonymous
  \author{Anonymous Authors}
\else
  \author{
  \authorhref{https://jishnujayakumar.github.io/}{Jishnu Jaykumar P},
  \authorhref{https://labs.utdallas.edu/irvl/people/}{Cole Salvato},
  \authorhref{https://labs.utdallas.edu/irvl/people/}{Vinaya Bomnale},
  \authorhref{https://jwroboticsvision.github.io/}{Jikai Wang}
  \\
  \authorhref{https://ayush-bhardwaj.vercel.app/}{Ayush Bhardwaj},
  \authorhref{https://jessekim.com/}{Jin Ryong Kim},
  \authorhref{https://yuxng.github.io/}{Yu Xiang}
  \\
  \rule{0pt}{1.6em}The University of Texas at Dallas
  }
\fi

\begin{document}

\makeatletter
\let\@oldmaketitle\@maketitle
\renewcommand{\@maketitle}{\@oldmaketitle
\centering
\vspace{1em}
\includegraphics[width=\linewidth]{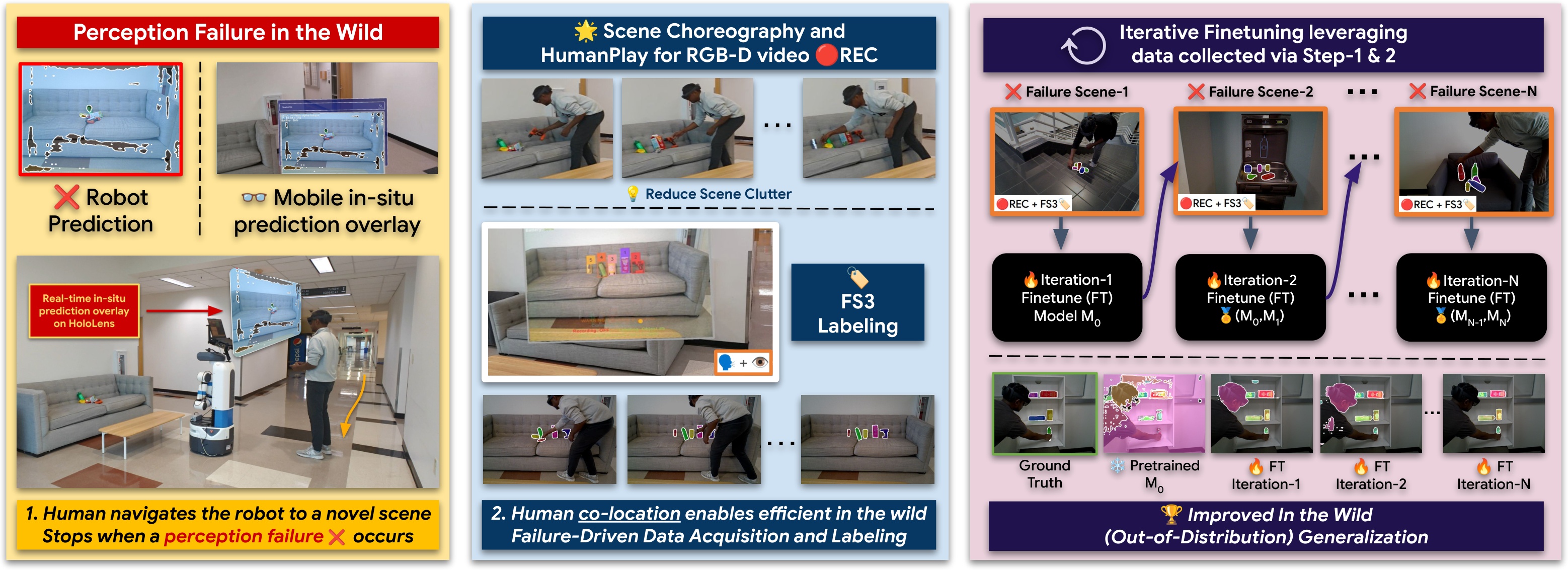}
\captionof{figure}{
Overview of {\coolname}. A pretrained perception (object segmentation) model fails under out-of-distribution 
real-world conditions due to clutter, occlusion, and novel objects (left). A co-located 
human wearing a mixed reality headset observes these failures and performs short 
human--object interaction (HumanPlay) to expose informative object configurations while 
recording an RGB-D sequence (middle). Only the final frame is annotated using hands-free 
eye-gaze and voice commands, and labels are propagated across the sequence by few-shot 
semi-supervised (FS3) labeling to generate dense supervision. The collected samples are used in a failure-driven 
iterative refinement loop (right).\ifanonymous\else\ Project page: \mbox{\url{\webpage}}\fi
\protect\phantomsection\label{fig:intro}
}
}
\vspace{-2.5em}
\makeatother

\maketitle

\vspace{-10em}
\begin{abstract}
We present \textit{iTeach}, a deployable system that lets any 
co-located human fix a robot's perception failures on the spot
without expertise, a workstation, or offline retraining. The operator wears 
a mixed reality (MR) headset, sees the robot's segmentation 
predictions overlaid on the real scene, and corrects failures 
hands-free: rearranging objects (HumanPlay), annotating via gaze 
and voice, and triggering SAM2 backward mask propagation. Each 
${\sim}$20\,s interaction yields 150--300 densely labeled training 
frames; the system fine-tunes the perception model onboard, keeps 
the better model, and redeploys, all without leaving the 
deployment site. The full loop requires only an RGB-D camera, onboard 
GPU, and an MR headset: any mobile robot, any environment. Starting from 26.1 on cluttered real-world scenes, 45 teaching interactions (13K frames) raise segmentation to 80.7 with no catastrophic forgetting; on three standard benchmarks the model never trained on, performance improves as well. Downstream 
pick-and-place on SceneReplica reaches 72/100, surpassing a 
model-based pipeline requiring CAD models. A 12-participant user study 
confirms non-experts match experts on annotation accuracy 
(${\sim}$95\% box IoU), speed, and task load (NASA-TLX 21/100). The 
framework is architecture-agnostic: any fine-tunable perception model 
can serve as backbone.
\end{abstract}


\setcounter{figure}{1}

\section{Introduction}
\label{sec:intro}

Robot perception models are trained on one distribution and deployed 
in another~\cite{khargonkar2024scenereplica, liu2024okrobot}. Training 
data (synthetic renders, curated benchmarks, controlled lab 
captures) cannot cover the clutter, objects, and lighting of every 
kitchen, warehouse, or hospital room the robot will encounter, and the 
resulting gap causes failures at every new site: missed objects, merged 
instances, spurious detections. Recent work increasingly treats such 
failures as a learning signal: VLMs diagnose manipulation failures 
from language~\cite{liu2023reflect, aha2024} and visual symbols guide 
corrective actions~\cite{zeng2026vifailback}. These methods operate at 
the \textit{action} level: they correct what the robot 
\textit{does} after a failure. Yet in modular pipelines where 
perception feeds planning and 
control~\cite{khargonkar2024scenereplica, liu2024okrobot, sundermeyer2021cgnet, xiang2024gto}, 
the root cause is often what the robot \textit{sees}: a missed object 
or merged mask cascades into a wrong grasp or a collision. Correcting 
the policy cannot help if the perception is wrong.

The standard perception fix is offline: collect site-specific data, 
annotate it at a workstation, retrain, and redeploy. But this cannot 
be failure-driven because the human cannot target failures they have not 
yet observed, and the scene may change before the retrained model 
ships. A failure-driven perception loop (see where the model fails 
now, collect targeted data there, retrain, and repeat) requires 
the human to be co-located with the robot, seeing predictions in 
real time. Any co-located human, not just an expert, should be able 
to see what the model gets wrong, correct it, and have the model 
improve on the spot.

We present {\coolname} (Fig.~\ref{fig:intro}), a system that realizes 
this loop using mixed reality. A human wearing a Microsoft HoloLens~2 
walks alongside the robot and sees its segmentation predictions overlaid 
on the physical scene. When the model fails, the human performs a short 
\textit{HumanPlay} interaction: physically transitioning the scene from 
cluttered to clean while the system records an RGB-D video (5--10\,s), 
then annotates only the clean final frame using eye-gaze and voice 
(1--3 point prompts per object). SAM2~\cite{ravi2024sam2} propagates 
those labels backward into the original cluttered frames, turning 
roughly 20 seconds of human effort into 150--300 labeled frames.

MR is what makes this loop practical. The headset is worn, not held, 
so hands stay free for HumanPlay while gaze and voice handle 
annotation. There is no screen to walk back to: the headset travels 
with the human across any environment and any mobile platform.

The labeled data feeds an iterative fine-tuning loop 
(Algorithm~\ref{alg:iteach}): fine-tune onboard, keep the better model, 
redeploy. Failure to redeployment takes 15--25 minutes, dominated by 
model-dependent fine-tuning time.

We evaluate {\coolname} on unseen object instance segmentation 
(UOIS)~\cite{XieCoRL19, xie2021unseen, msmformer} with 
MSMFormer~\cite{msmformer} as the base model, but the principle 
generalizes: any fine-tunable perception model can serve as backbone, 
and the same failure-driven loop extends to other perception tasks 
and, ultimately, to manipulation itself. Our contributions:

\begin{enumerate}
\item \textbf{Deployable system.} A complete system for correcting 
perception failures during deployment: observe through MR, collect via 
hands-free interaction, retrain onboard, redeploy, with no workstation 
and no expert. It needs only an RGB-D camera, an onboard GPU, and an 
MR headset. A 12-participant user study confirms non-experts match 
experts on annotation accuracy ({$\sim$}95\% IoU), speed, and task load.

\item \textbf{Data-efficient teaching mechanism.} HumanPlay transitions 
scenes from cluttered to clean while recording video; the human 
annotates one frame via gaze and voice; SAM2 propagates labels 
backward into the cluttered frames. Each {$\sim$}20\,s interaction 
yields 150--300 densely labeled frames, three orders of magnitude 
more efficient than manual per-frame annotation.

\item \textbf{End-to-end validation from perception to manipulation.} 
45~teaching interactions (13K~frames) raise UOIS from 26.1 to 80.7 
on cluttered real-world scenes with no degradation on three standard 
benchmarks. The perception gains propagate to manipulation: 
pick-and-place success on 
SceneReplica~\cite{khargonkar2024scenereplica} reaches 72/100, 
matching a model-based pipeline that requires known CAD models. The 
framework is architecture-agnostic: any fine-tunable perception model 
can replace MSMFormer.
\end{enumerate}

\section{Related Work}\label{sec:related_work}

\textbf{Human-in-the-loop robot learning.}
Most human-in-the-loop methods operate at the action 
level~\cite{ross2011reduction, kelly2019hgdagger, hoque2021thriftydagger, celemin2022interactive}: 
a human corrects trajectories or provides reward signals to refine a 
policy~\cite{liu2023sirius, liu2024olaf, luo2024hilserl}. In modular 
systems~\cite{allu2025modularroboticautonomousexploration, lu2024nidsnet, liu2024okrobot, khargonkar2024scenereplica}, 
however, failures often originate in perception rather than control. 
{\coolname} targets this gap: the human corrects the perception model, 
not the policy.

\textbf{Data collection for robot perception.}
Interactive perception~\cite{bohg2017interactive} uses robot actions to 
generate informative training signal. Lu~et~al.~\cite{lu2023sss} 
demonstrated that a robot can autonomously push objects apart to collect 
self-supervised segmentation data. While effective in tabletop settings, 
robot-driven exploration cannot reliably target perception failures or 
operate beyond structured environments. Large-scale teleoperated data 
collection~\cite{zhao2023aloha, mandlekar2019roboturk, kareer2024egomimic, chi2024umi} 
and label-efficient 
methods~\cite{rise2024, zeng2020transporter, padalunkal2024protoclip} address scale but operate 
offline. {\coolname} collects data \textit{during deployment}, at the 
point of failure, with minimal human effort.

\textbf{Real-world adaptation.}
Sim-to-real approaches include dataset 
diversity~\cite{khazatsky2024droid, openxembodiment2024}, domain 
randomization~\cite{tobin2017domain}, and test-time adaptation (TTA). 
TTA adjusts on unlabeled target data but cannot fix structural errors 
(missed objects, merged instances) that need new labeled examples; 
active learning selects from an existing pool, whereas {\coolname} 
\textit{creates} new data at the point of failure. 
Schreiber~et~al.~\cite{schreiber2025doyouknow} correct traversability 
in the field in seconds; our setting is harder (dense masks, larger 
model) but shares the failure-driven, in-situ philosophy.

\textbf{Mixed reality for robotics.}
Prior MR work in robotics focuses on visualization of robot 
intent~\cite{rosen2020communicating}, 
programming and demonstration 
collection~\cite{arunachalam2023holo}, and offline wearable data 
capture~\cite{liu2025egozero, kareer2024egomimic}. 
Bohus~et~al.~\cite{bohus2022continual} use an MR headset for 
continual object recognition via speech, gaze, and gestures, but 
target classification (closed-set labeling) rather than dense instance 
segmentation, and do not close the loop through a robot. We use MR as 
a \textit{teaching interface}: the headset overlays predictions on the 
real scene so the human can diagnose failures, and its hands-free form 
factor allows simultaneous object manipulation and annotation,
which a screen or tablet cannot provide together.

\begin{figure}[!t]
    \centering
    \includegraphics[width=\linewidth]{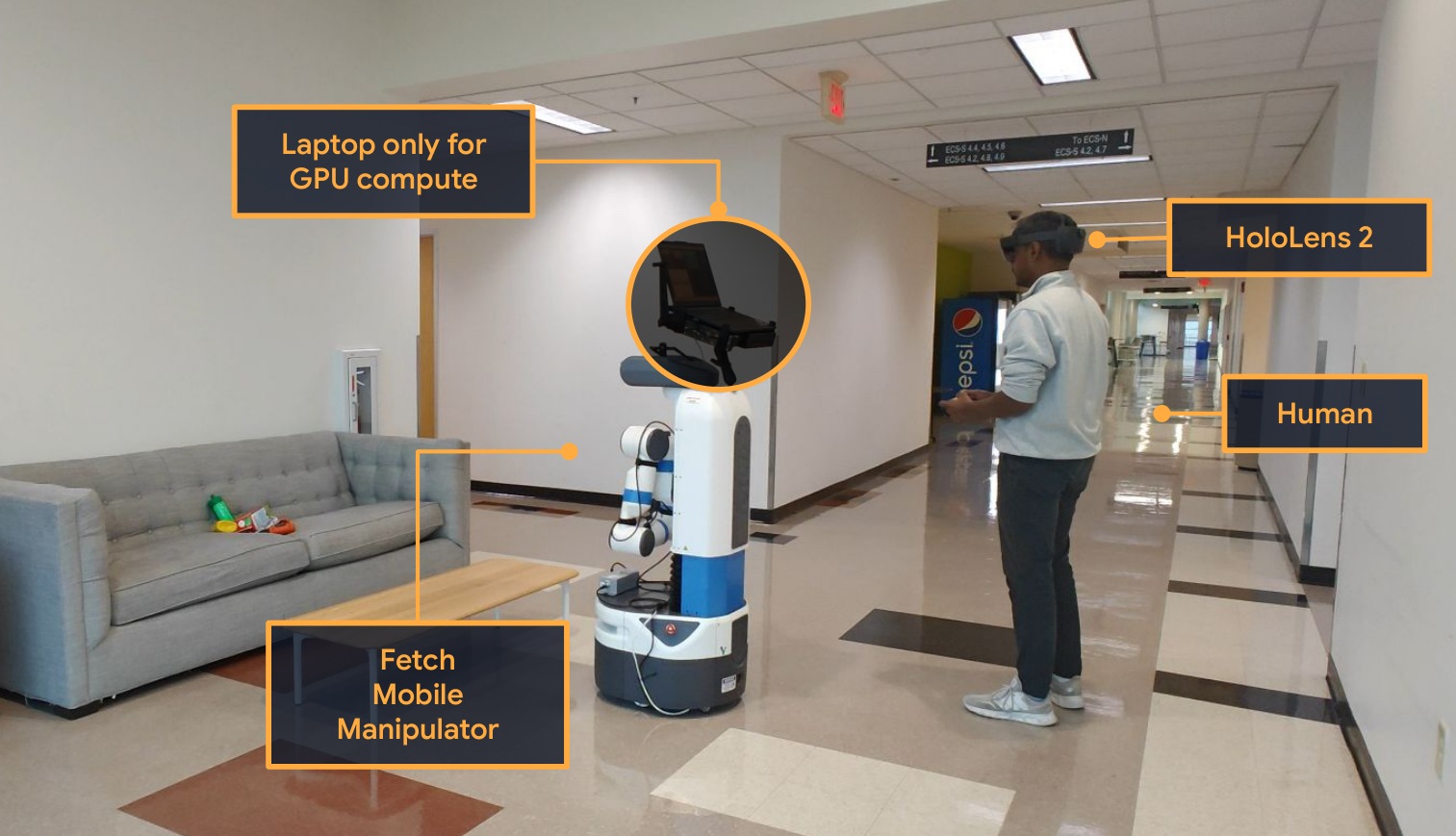}
    \caption{System setup. A Fetch robot streams RGB-D to an onboard 
    GPU (RTX~4090 in our prototype; replaceable by embeddable 
    compute such as Jetson Thor). A HoloLens~2 connected via Wi-Fi 
    overlays predictions on the real scene and captures gaze-and-voice 
    annotations. All inference, SAM2 propagation, and fine-tuning run 
    onboard.}
    \label{fig:sys-components}
    \vspace{-4mm}
\end{figure}

\section{Method}
\label{sec:method}

\subsection{System Overview}

The system has two logical components 
(Fig.~\ref{fig:sys-components}): (1)~any mobile robot with an 
RGB-D camera and onboard GPU, and (2)~a mixed reality headset worn by 
the human. We demonstrate on a Fetch mobile manipulator carrying an 
RTX~4090 module, but the architecture is robot-agnostic: a 
humanoid, a quadruped, or any mobile platform with RGB-D sensing and 
sufficient compute can serve as the robot side; an embeddable module 
such as NVIDIA Jetson Thor would make the system fully self-contained. 
The onboard computer is purely infrastructure: it runs inference, 
streams predictions to the headset, receives annotations, runs SAM2 
propagation, and fine-tunes. The human never interacts with it 
directly; no cloud or external compute is required.

\subsection{Failure Observation}

\begin{figure}[h]
    \centering
    \includegraphics[width=\linewidth]{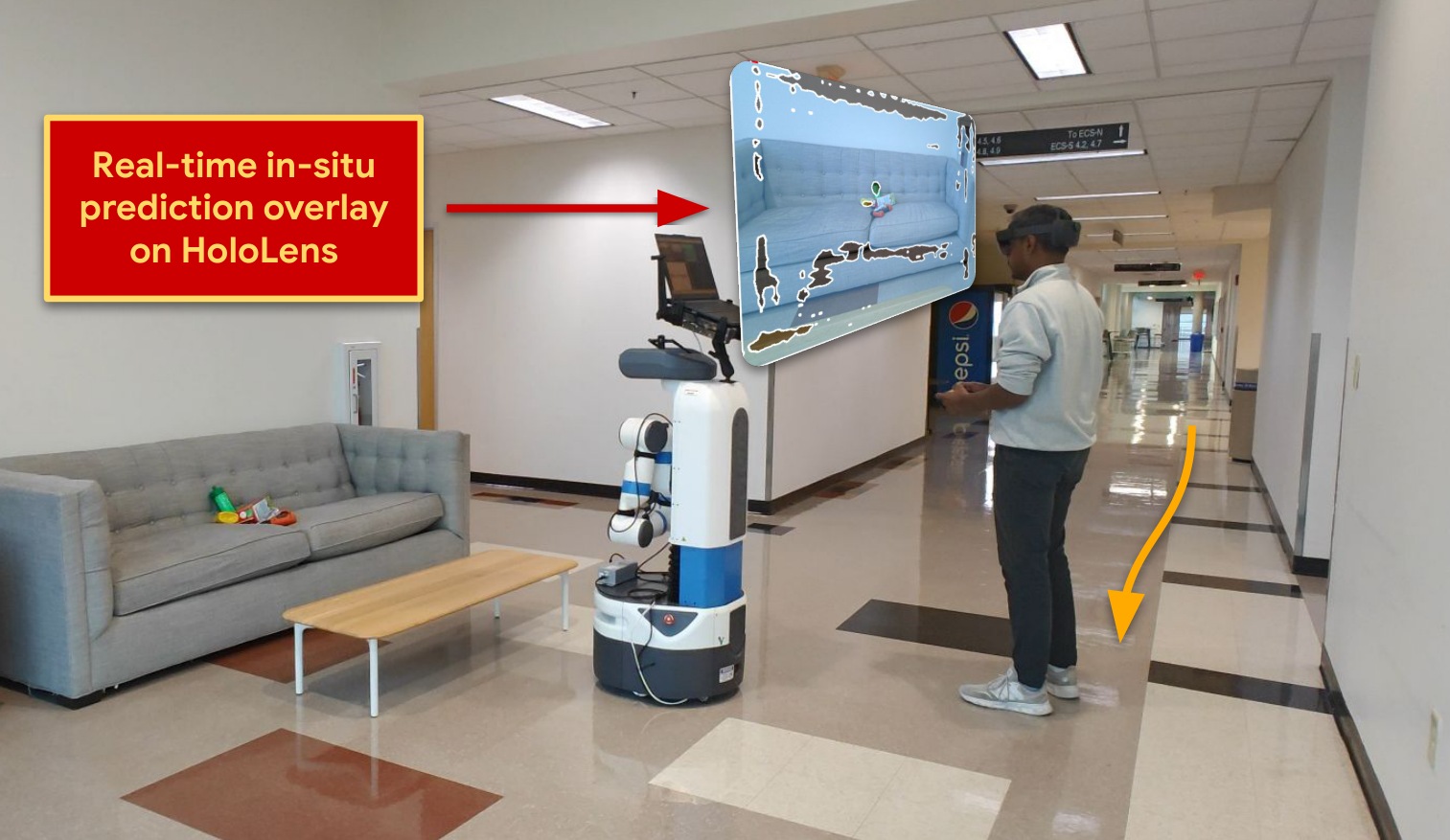}
    \caption{The human navigates the robot through cluttered environments. 
    When the MR overlay reveals a segmentation failure (missed objects, 
    merged instances, or spurious masks), the human initiates data 
    collection.}
    \label{fig:task-space-setup}
\end{figure}

The human walks with the robot and sees its segmentation predictions 
rendered in the MR headset, overlaid on the physical scene 
(Fig.~\ref{fig:task-space-setup}). A perception failure is any case 
where predicted masks visibly deviate from actual object boundaries: 
missed objects, merged instances, spurious detections, or grossly 
inaccurate boundaries. The human steers the robot toward cluttered 
scenes of diverse everyday objects~\cite{padalunkal2023fewsol}, where 
failures are likelier.

\begin{figure*}[!t]
    \centering
    \includegraphics[width=.85\linewidth]{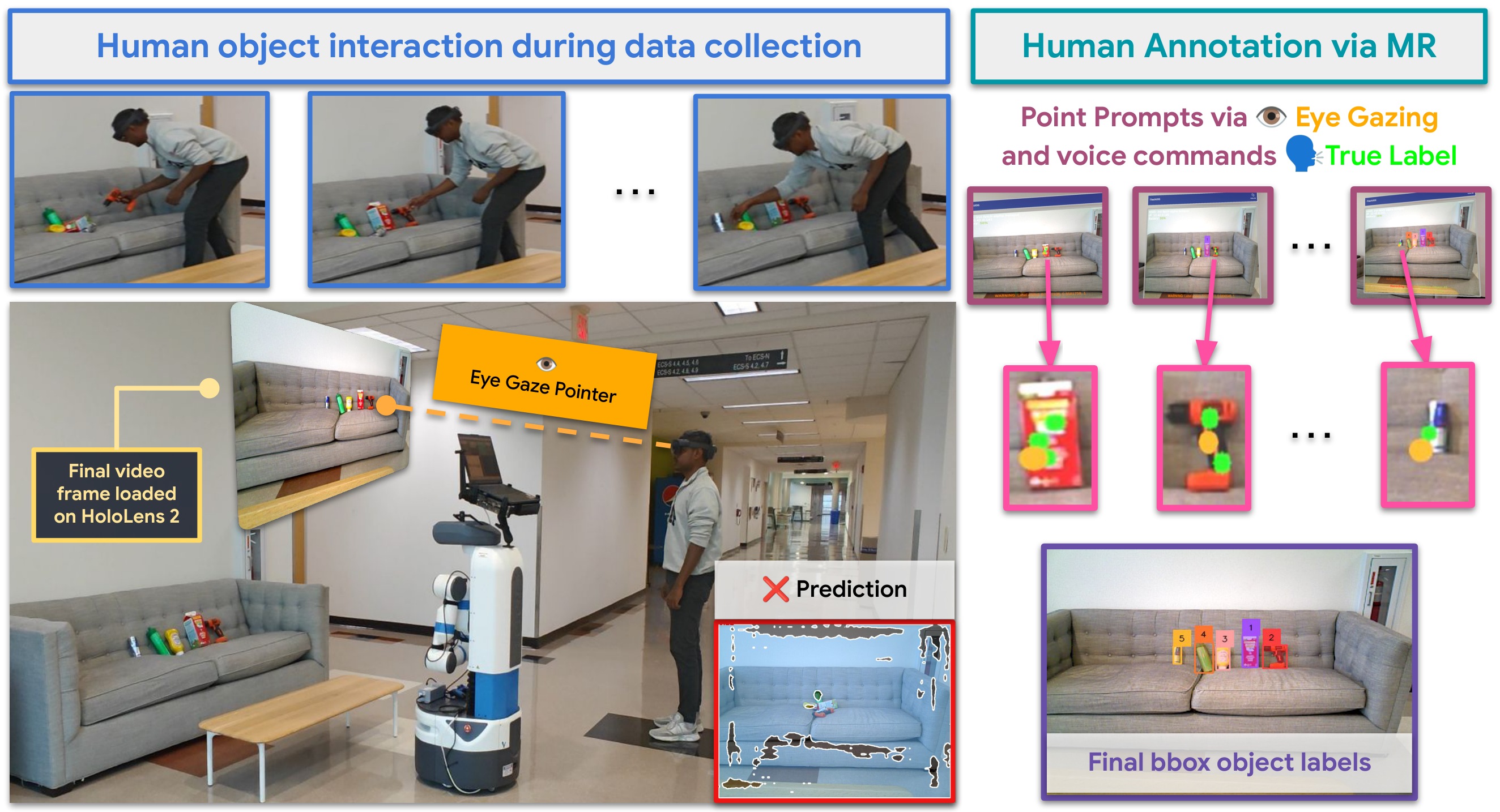}
    \caption{Annotation via MR. The orange cursor tracks eye gaze; green 
    dots are point prompts placed by voice command. SAM2~\cite{ravi2024sam2} 
    converts point prompts to bounding boxes. HumanPlay transitions the 
    scene from cluttered to clean, enabling accurate final-frame annotation.}
    \label{fig:iteach-uois-annotation}
    \vspace{-5mm}
\end{figure*}

\begin{figure}[!t]
    \centering
    \includegraphics[width=\linewidth]{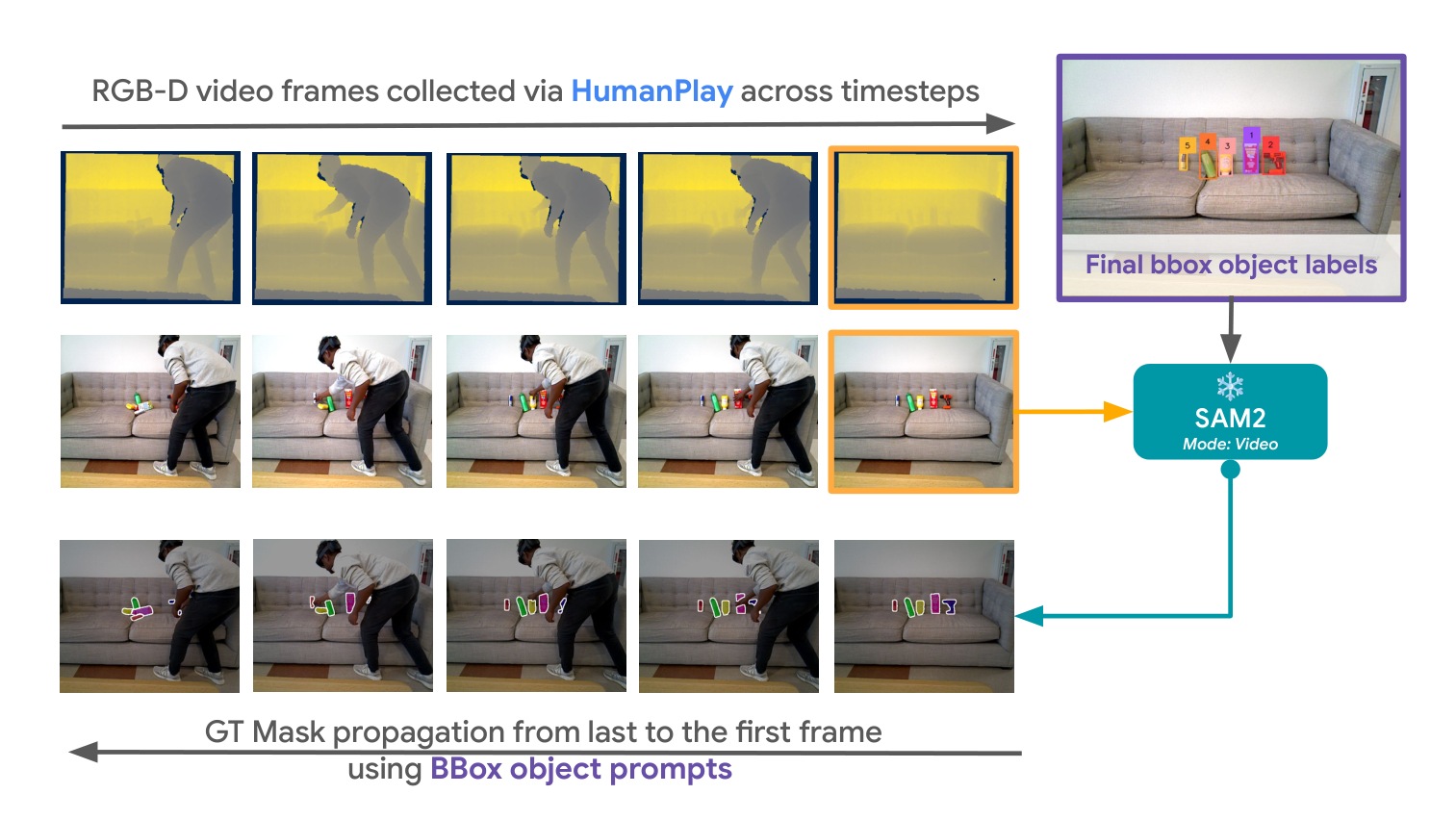}
    \caption{SAM2 backward propagation. Masks generated from the 
    annotated final frame are propagated to earlier, more cluttered 
    frames, converting one annotation into dense video-level 
    supervision.}
    \label{fig:sam2-mask-prop}
    \vspace{-5mm}
\end{figure}

\subsection{HumanPlay and Annotation}
\label{subsec:human-interaction-and-annotation}

When a failure is spotted, the human performs \textit{HumanPlay}: 
physically rearranging objects to reduce occlusion while the system 
records an RGB-D video (5--10\,s); the robot arm remains stationary 
throughout. The result is a short sequence in 
which the scene transitions from cluttered (where perception failed) to 
separated (where individual objects are clearly visible). The human then 
annotates only the final frame using eye-gaze and voice commands 
(Fig.~\ref{fig:iteach-uois-annotation}), typically placing 1--3 point 
prompts per object in 5--10\,s per object.

SAM2~\cite{ravi2024sam2} converts the point prompts to bounding-box 
proposals (image mode), then propagates masks backward through the 
recorded video (video mode), producing 
dense per-frame segmentation labels 
(Fig.~\ref{fig:sam2-mask-prop}). A single HumanPlay interaction thus 
yields 150--300 labeled training frames from one annotation act. 
We refer to this as few-shot semi-supervised (FS3) labeling.

\textbf{Why HumanPlay and video, not single-frame annotation?}
The alternative is to annotate the cluttered failure frame directly, 
where objects overlap and SAM2 has no temporal context. HumanPlay 
gives two advantages: (1)~separated objects make prompts more accurate 
and masks cleaner, and (2)~backward propagation yields $N$ labeled 
frames instead of one, and the early frames \textit{are} the 
cluttered scenes where perception failed 
(Sec.~\ref{subsec:uois-results}). Point prompts need 
only land on the target object; HoloLens~2 gaze accuracy 
($\sim$1.5$^\circ$) suffices.

\textbf{What is novel vs.\ what is off-the-shelf.}
SAM2 is an off-the-shelf tool. The contribution is the workflow it 
sits inside: failure-driven scene selection, HumanPlay, sparse MR 
supervision, and iterative model improvement.

\begin{figure*}[!t]
\centering
\includegraphics[width=.93\linewidth]{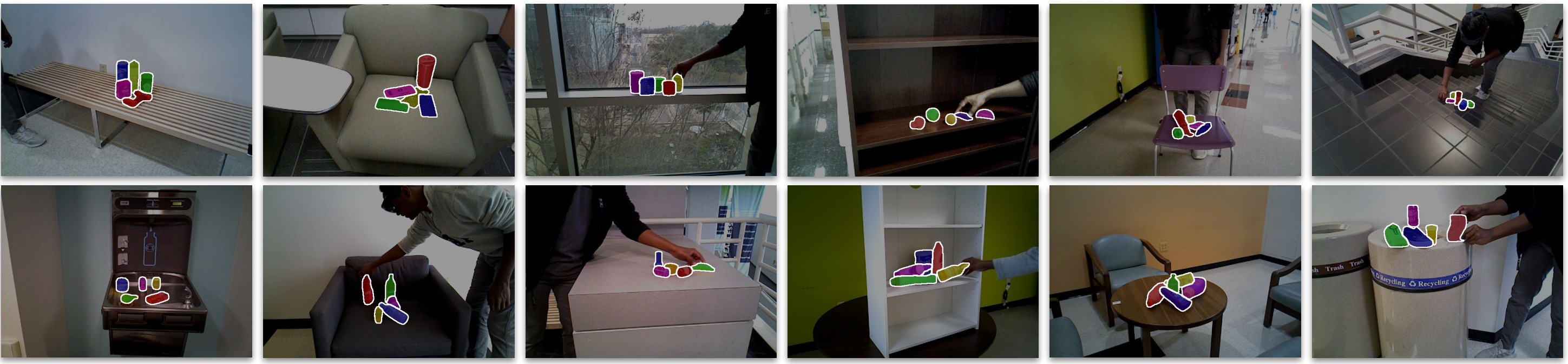}
\caption{Annotated data collected via MR across diverse environments.}
\label{fig:hololens-annotated}
\vspace{-5mm}
\end{figure*}

\subsection{Iterative Fine-Tuning}
\label{subsec:iterative-ft}

\begin{algorithm}[t]
\caption{{\coolname}: Iterative Failure-Driven Adaptation}
\label{alg:iteach}
\begin{algorithmic}[1]
\Require Pretrained perception model $M_0$
\Ensure Adapted perception model $M_t$
\State $\mathcal{D} \leftarrow \emptyset$,\; $t \leftarrow 0$
\While{deployment is active}
    \State Deploy $M_t$; run inference on RGB-D stream
    \State Render predictions in MR overlay
    \If{perception failure observed}
        \State $V \leftarrow \textsc{RecordVideo}(\text{HumanPlay interaction})$ \Comment{$T$ frames}
        \State $\mathbf{p} \leftarrow \textsc{AnnotateFrame}(V_T)$ \Comment{gaze $+$ voice}
        \State $\mathbf{b} \leftarrow \textsc{SAM2}_{\text{image}}(V_T,\, \mathbf{p})$ \Comment{point $\to$ box}
        \State $\mathcal{D}_{\text{new}} \leftarrow \textsc{SAM2}_{\text{video}}(V,\, \mathbf{b})$ \Comment{propagate masks}
        \State $\mathcal{D} \leftarrow \mathcal{D} \cup \mathcal{D}_{\text{new}}$
        \State $\tilde{M}_t \leftarrow \textsc{FineTune}(M_t,\, \mathcal{D})$
        \State $M_{t+1} \leftarrow \arg\max_{M \in \{M_t,\, \tilde{M}_t\}} C(M)$
        \State $t \leftarrow t + 1$
    \EndIf
\EndWhile
\end{algorithmic}
\end{algorithm}

The collected data feeds an iterative fine-tuning loop 
(Algorithm~\ref{alg:iteach}). After each teaching interaction, the 
system fine-tunes the current model on the accumulated dataset and 
compares the result against the deployed model using the combined 
score $C$ (defined below) on held-out test scenes, keeping the better 
one. This best-of-two selection prevents degradation: the model 
improves or stays the same, never gets worse. Without labeled test 
data the human makes this call visually via the MR overlay. There is no 
formal stopping criterion; the loop runs as long as the human 
observes failures.

\subsection{Evaluation Metrics}
\label{subsec:metrics}

We adopt standard UOIS metrics~\cite{msmformer, lu2023sss}: 
overlap precision $P$, recall $R$, and F-measure $F$:
\begin{gather*}
P = \frac{\sum_i |\hat{M}_i \cap M_{\pi(i)}|}{\sum_i |\hat{M}_i|},\quad
R = \frac{\sum_i |\hat{M}_i \cap M_{\pi(i)}|}{\sum_j |M_j|},\\
F = \frac{2PR}{P+R},
\end{gather*}
where $\hat{M}_i$ is the $i$-th predicted mask, $M_j$ the $j$-th 
ground-truth mask, and $\pi(i)$ the optimal assignment via Hungarian 
matching. We report object F-measure ($F_o$), boundary F-measure 
($F_b$), and $I_{0.75}$ (fraction with $F_o \ge 0.75$), capturing 
object, boundary, and IoU-level performance respectively. The composite score is:
\[
C = 0.4 \cdot F_o + 0.4 \cdot F_b + 0.2 \cdot I_{0.75}.
\]

\section{Experiments}\label{sec:experiments}

We evaluate the system end-to-end: (1)~Can it recover from 
perception failures? (2)~Does recovery generalize without forgetting? 
(3)~Do perception gains translate to manipulation? (4)~Can 
non-experts operate it?

\subsection{Setup}\label{subsec:setup}

\textbf{Base model.}
We use MSMFormer~\cite{msmformer}, a two-stage UOIS model pretrained on 
synthetic tabletop data~\cite{XieCoRL19}. It performs well on standard 
benchmarks (OCID~\cite{ocid}, OSD~\cite{osd}, 
RobotPushing~\cite{lu2023sss}) but scores only 26.1 on cluttered 
real-world scenes, the sim-to-real gap typical of synthetic-only 
training. We choose MSMFormer over recent alternatives such as 
RISeg~\cite{qian2024riseg}, rt-RISeg~\cite{qian2025rtriseg}, and 
ZISVFM~\cite{zhang2025zisvfm} for three reasons: 
(1)~its code and pretrained weights are publicly available (the 
alternatives above lack public weights at the time of writing), 
enabling reproducibility; 
(2)~Lu~et~al.~\cite{lu2023sss} previously fine-tuned the same model 
with robot-collected data, giving a direct, controlled comparison of 
data sources (their autonomous collection vs.\ our failure-driven 
collection); and 
(3)~it is the perception backbone in the 
SceneReplica~\cite{khargonkar2024scenereplica} manipulation pipeline, 
letting us measure whether perception gains propagate to task success 
end-to-end. 
The {\coolname} approach is not specific to MSMFormer; any fine-tunable 
UOIS model could serve as backbone. We use MSMFormer to isolate the 
effect of \textit{how data is collected}, not which architecture 
processes it.

\textbf{Dataset.}
Using {\coolname}, we collected 48 real-world scenes 
(45 train, 3 test) spanning tabletops, sofas, shelves, stairs, and 
floor-level arrangements (Fig.~\ref{fig:hololens-annotated}), each a 
5--10\,s HumanPlay video of diverse household and office objects. SAM2 
propagation succeeded in 35/40 training scenes (87.5\%); failures 
involved heavy occlusion, a SAM2 limitation, not a framework 
limitation. The dataset contains 
${\sim}$13,000 training frames and 902 test frames from 
45 interactions of $\sim$20\,s human effort each. Test-set ground 
truth was annotated on a desktop (not via MR), so evaluation is 
unbiased.

\begin{table}[!t]
\centering
\resizebox{\linewidth}{!}{
\begin{tabular}{c|c|ccc|c}
\hline
\textbf{Round} & \textbf{\#Scenes} & \cellcolor{blue!10}$\mathbf{F_o}$ & \cellcolor{blue!10}$\mathbf{F_b}$ & \cellcolor{blue!10}$\mathbf{I_{0.75}}$ & \cellcolor{orange!15}$\mathbf{C}$ \\
\hline
\rowcolor{gray!12}
Pre. & -- & 28.2 & 21.7 & 30.9 & 26.1 \\
\hline
1 & 3  & 69.0 & 64.5 & 75.0 & 68.4 \\
2 & 6  & 70.5 & 66.2 & 74.1 & 69.5 \\
3 & 9  & 78.3 & 72.3 & 77.4 & 75.7 \\
4 & 12 & 82.2 & 74.4 & 79.5 & 78.5 \\
5 & 15 & 74.4 & 68.0 & 78.9 & 72.7 \\
\rowcolor{green!10}
6 & 20 & \textbf{82.8} & \textbf{76.0} & 83.7 & 80.3 \\
\rowcolor{green!8}
8 & 30 & 82.7 & 75.6 & \textbf{86.7} & \textbf{80.7} \\
\hline
\end{tabular}}
\caption{Iterative fine-tuning on D40 (902 test images). The pretrained 
model ($C=26.1$) is trained on synthetic data only. Performance saturates 
at $\sim$20 scenes. D5 (not shown) reaches $C=76.6$ after 5 single-scene 
rounds.}
\label{tbl:iteach-uois-stage-1-ablation}
\end{table}

\begin{figure}[!t]
    \centering
    \includegraphics[width=.92\linewidth]{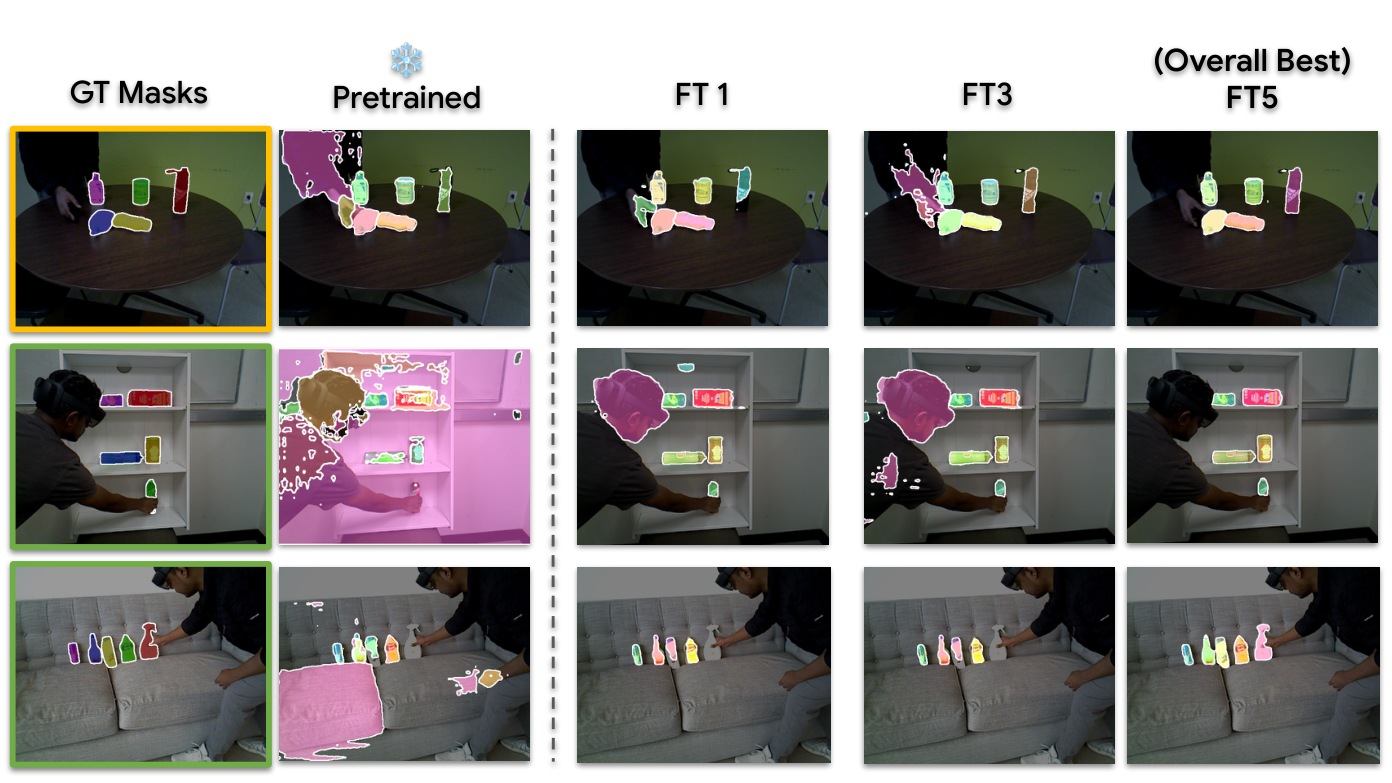}
    \caption{Qualitative UOIS results across fine-tuning stages on 
    tabletop and beyond-tabletop scenes.}
    \label{fig:iteach-qual-comparison}
    \vspace{-5mm}
\end{figure}

\subsection{Recovery from Perception Failures}\label{subsec:uois-results}

We fine-tune Stage~1 of MSMFormer on a synthetic--real mixed dataset 
from pretrained weights. Each round follows Algorithm~\ref{alg:iteach}: 
collect failure-driven data from the current model, fine-tune, keep the 
better model.

\textbf{D5 (one scene per round).}
Adding a single scene per round improves $C$ from 26.1 to 76.6 after 5 
rounds ($\Delta C = +50.5$); one well-targeted interaction alone takes 
it from 26.1 to 62.8 in round~1. This demonstrates that even a single 
failure-driven teaching episode (roughly 20\,s of human effort) 
can recover nearly half the sim-to-real gap, because the human 
naturally targets scenes where the model fails most.

\textbf{D40 (multiple scenes per round).}
Table~\ref{tbl:iteach-uois-stage-1-ablation} shows the scaling 
behavior: performance rises sharply in early rounds and saturates 
around 20~scenes ($C = 80.3$, $\Delta C = +54.2$), peaking at 80.7 with 
30~scenes. Beyond 20~scenes returns are marginal; scene diversity 
matters more than sheer volume, echoing the saturation effect 
reported by Lu~et~al.~\cite{lu2023sss} for robot-collected data.

Fig.~\ref{fig:iteach-qual-comparison} shows qualitative improvements 
and Table~\ref{tbl:iteach-uois-stage-1-ablation} reports the full 
scaling curve.

\subsection{Generalization Without Forgetting}
\label{subsec:comparison}

A natural concern is catastrophic forgetting: does fine-tuning on 
in-the-wild data degrade the benchmarks MSMFormer was designed for? 
Table~\ref{tbl:longer-training} compares against the pretrained 
baseline and against Lu~et~al.~\cite{lu2023sss}, who fine-tune 
MSMFormer with self-supervised robot-pushing data. To isolate training 
data from the iterative schedule, we train 50K iterations (vs.\ 2K in 
the loop) from the same initialization on aggregated HumanPlay data.

\begin{table}[!t]
\centering
\resizebox{\linewidth}{!}{
\begin{tabular}{l|c|c|c|c}
\hline
\textbf{Model} & \textbf{OCID}~\cite{ocid} & \textbf{OSD}~\cite{osd} & \textbf{RobotPush}~\cite{lu2023sss} & \textbf{iTeach-HP} \\
 & $C$ & $C$ & $C$ & $C$ \\
\hline
\rowcolor{gray!12}
MSMFormer~\cite{msmformer} & 86.7 & 76.4 & 83.4 & 29.3 \\
Lu et al. (OCID)~\cite{lu2023sss} & 87.1 & 74.6 & 85.6 & 26.9 \\
Lu et al. (OSD)~\cite{lu2023sss}  & 87.4 & 76.2 & 89.0 & 39.9 \\
iTeach-UOIS$^\dagger$ (scratch)   & 76.5 & 69.7 & 80.6 & 58.3 \\
\rowcolor{green!10}
\textbf{iTeach-UOIS$^\ddagger$} (pretrained init) & \textbf{87.6} & \textbf{77.5} & \textbf{87.8} & \textbf{67.8} \\
\hline
\end{tabular}}
\caption{Longer training (50K iterations) on aggregated HumanPlay data. 
iTeach-UOIS$^\dagger$: HumanPlay data only, no pretraining. 
iTeach-UOIS$^\ddagger$: pretrained MSMFormer + HumanPlay fine-tuning. 
The failure-driven data improves the target domain (+38.5 on iTeach-HP) 
while maintaining OCID (+0.9), OSD (+1.1), and RobotPushing (+4.4).}
\label{tbl:longer-training}
\end{table}

iTeach-UOIS$^\ddagger$ improves on \textit{all four} datasets, 
including benchmarks it never trained on (OCID +0.9, OSD +1.1, 
RobotPushing +4.4), with no catastrophic forgetting. The improvement 
on RobotPushing is notable: the model has never seen robot-pushing 
data, yet the diversity of HumanPlay scenes (which span tabletops, 
sofas, shelves, and floor-level arrangements) transfers to this 
distribution as well. Meanwhile, 
Lu~et~al.'s robot-pushing data provides limited transfer to in-the-wild 
scenes ($\Delta C = -2.4$ for OCID-collected, $+10.6$ for OSD-collected), 
far below the $+38.5$ from failure-driven collection, confirming that 
\textit{how} data is selected matters more than how much is collected.

\begin{figure*}[!t]
    \centering
    \includegraphics[width=.98\linewidth]{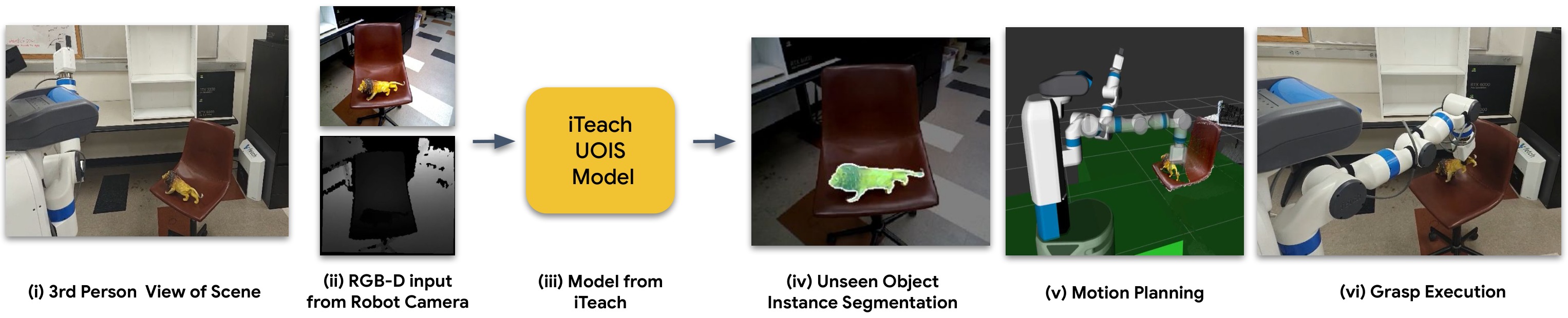}
    \caption{Pick-and-place execution using Pipeline~4. {\coolname}-UOIS 
    produces clean instance masks, enabling accurate grasp planning 
    and collision-free motion.}
    \label{fig:rw-manipulation}
    \vspace{-5mm}
\end{figure*}

\subsection{From Perception to Manipulation}\label{subsec:manipulation}

We swap the perception module in a 
SceneReplica~\cite{khargonkar2024scenereplica} pipeline and keep 
everything else identical (same grasp planner 
(Contact-GraspNet~\cite{sundermeyer2021cgnet}), motion planner 
(GTO~\cite{xiang2024gto}), and control (MoveIt)), so any change in 
success rate is attributable to perception alone.

\begingroup
\renewcommand{\arraystretch}{1.3}
\begin{table}[!t]
\centering
\resizebox{\linewidth}{!}{
\begin{tabular}{c|l|l|c|c}
\hline
\textbf{\#} & \textbf{Perception} & \textbf{Motion} & \textbf{Grasp $\uparrow$} & \textbf{P\&P $\uparrow$} \\
\hline
1 & GDRNPP~\cite{Liu_gdrnpp} (model-based) & RFP+OMPL~\cite{sucan2012OMPL} & 73 & 70 \\
2 & MSMFormer~\cite{msmformer} & CGNet+OMPL & 65 & 57 \\
3 & MSMFormer~\cite{msmformer} & CGNet+GTO & 71 & 65 \\
\rowcolor{green!10}
4 & \textbf{{\coolname}-UOIS} & CGNet+GTO & \textbf{74} & \textbf{72} \\
\hline
\end{tabular}}
\caption{SceneReplica manipulation results (out of 100 trials). 
Pipeline~4 differs from Pipeline~3 only in perception. Pipeline~1 uses 
model-based perception (GDRNPP) with known CAD models. 
{\coolname}-UOIS achieves the best result on this benchmark, 
surpassing model-based perception (GDRNPP) without requiring CAD models.}
\label{tab:scene-replica}
\end{table}
\endgroup

Pipeline~4 ({\coolname}-UOIS) improves over Pipeline~3 (pretrained 
MSMFormer) by +3 grasps and +7 pick-and-place successes 
(Table~\ref{tab:scene-replica}), matching the model-based Pipeline~1 
that requires known CAD models, without any object models at all 
(Fig.~\ref{fig:rw-manipulation}). The improvement is entirely due to 
perception: cleaner instance masks produce tighter point clouds for 
grasp planning, reducing collisions and missed grasps. This validates 
the cascade hypothesis from Sec.~\ref{sec:intro}: fixing what the 
robot \textit{sees} directly fixes what it \textit{does}, even without 
touching the policy.

\subsection{System Latency}\label{subsec:timing}

\begin{table}[!t]
\centering
\begin{tabular}{l|r|l}
\hline
\textbf{Stage} & \textbf{Duration} & \textbf{Details} \\
\hline
\rowcolor{green!10}
HumanPlay & 5--10\,s & Rearrange objects \\
\rowcolor{green!10}
Annotation & 5--10\,s/obj & Eye-gaze + voice \\
\hline
\rowcolor{orange!15}
SAM2 propagation$^\dagger$ & $\sim$6\,FPS & RTX 4090 \\
\rowcolor{orange!15}
Dataset update$^\dagger$ & $<$2\,min & Aggregate frames \\
\rowcolor{orange!15}
Fine-tuning$^\dagger$ & 10--15\,min & 2K iters, mixed dataset \\
\hline
\rowcolor{orange!15}
\textbf{Total (incl.\ model)} & \textbf{15--25\,min} & Failure $\rightarrow$ redeployment \\
\hline
\end{tabular}
\caption{Timing breakdown. \colorbox{green!10}{Green}: 
system-constant ($\sim$20\,s human effort per scene). 
\colorbox{orange!15}{Orange}$^\dagger$: model-dependent (SAM2 + 
MSMFormer on RTX~4090); reducible with faster models or LoRA.}
\label{tbl:timing}
\vspace{-6mm}
\end{table}

Table~\ref{tbl:timing} breaks down the loop: human effort is 
$\sim$20\,s per scene (system-constant); the remaining time is 
model-dependent computation, dominated by fine-tuning 
(10--15\,min with MSMFormer at 48.9\,M params). Lighter backbones 
such as rt-RISeg~\cite{qian2025rtriseg} ($\sim$3\,M params) or 
parameter-efficient methods (LoRA, $\sim$1\,M trainable) could bring 
the cycle under 10\,minutes. Predicted masks are streamed to the 
HoloLens~2 with sub-second end-to-end latency.

\subsection{System Design: Why Mixed Reality?}\label{subsec:why-mr}

The loop requires simultaneously seeing predictions, rearranging 
objects, and annotating, all hands-free, co-located, in any 
environment. No desktop or tablet satisfies all three 
(Table~\ref{tbl:interface}). MR is the enabling technology: the 
headset is worn, gaze and voice handle annotation, hands handle 
objects, and the system travels with the human.

\begin{table}[!t]
\centering
\resizebox{\linewidth}{!}{
\begin{tabular}{l|c|c|c}
\hline
\textbf{Capability} & \textbf{MR (ours)} & \textbf{Desktop} & \textbf{Tablet} \\
\hline
Works across mobile robots & {\color{green!60!black}\textbf{\checkmark}} & {\color{red}\textbf{$\times$}} & {\color{red}\textbf{$\times$}} \\
Co-located with robot    & {\color{green!60!black}\textbf{\checkmark}} & {\color{red}\textbf{$\times$}}   & {\color{green!60!black}\textbf{\checkmark}} \\
In-situ prediction overlay & {\color{green!60!black}\textbf{\checkmark}} & {\color{red}\textbf{$\times$}} & {\color{red}\textbf{$\times$}} \\
Hands-free object interaction\textsuperscript{*} & {\color{green!60!black}\textbf{\checkmark}} & {\color{green!60!black}\textbf{\checkmark}} & {\color{red}\textbf{$\times$}} \\
Hands-free annotation    & {\color{green!60!black}\textbf{\checkmark}} & {\color{red}\textbf{$\times$}}   & {\color{red}\textbf{$\times$}} \\
\hline
All five simultaneously  & {\color{green!60!black}\textbf{\checkmark}} & {\color{red}\textbf{$\times$}}   & {\color{red}\textbf{$\times$}} \\
\hline
\end{tabular}}
\caption{ Interface comparison. A desktop cannot travel with the robot; a tablet can, but must be held and shows the scene on a screen rather than in it. Only a worn headset delivers all five capabilities the \cooltitle loop needs at once. *A held tablet occupies one hand, so neither guiding the robot nor rearranging objects stays hands-free.
\vspace{-5mm}}
\label{tbl:interface}
\end{table}
\subsection{MR User Study}\label{subsec:user-study}

The argument above is a design argument; we test it with a user study 
that asks a narrower question: are the bounding boxes produced by 
gaze-and-voice annotation accurate enough, and can any user produce 
them after minimal training? The study does \textit{not} evaluate 
downstream model training; all training data in this paper was 
collected by one experienced operator. We recruited 
12~participants: 6~\textit{experts} (prior robotics or MR 
experience) and 6~\textit{non-experts}.
\ifanonymous
\else
The study was conducted under UTD IRB protocol 21-194 (PI: Jin Ryong 
Kim); all participants gave informed consent.%
\fi

\textbf{Protocol.} After a short briefing, participants did practice trials until they felt comfortable. Each participant then completed one full teaching interaction unassisted: observe the failure through the headset, perform HumanPlay, annotate the final frame with gaze and voice, covering 20 objects (240 annotations total). \textit{Labeling accuracy} is the mean IoU between the box SAM2 
derives from the participant's point prompts and a reference box. 
\textit{Completion time} sums HumanPlay and annotation time. 
\textit{Task load} is NASA-TLX~\cite{hart1988nasatlx}, administered 
once per session as the unweighted mean of its six subscales, every 
scale oriented so that lower is better. Figures report group means with 
error bars of one standard error of the mean, SEM~$=$~SD$/\sqrt{6}$, 
where SD is the standard deviation.

\begin{figure}[!t]
\centering
\includegraphics[width=\linewidth]{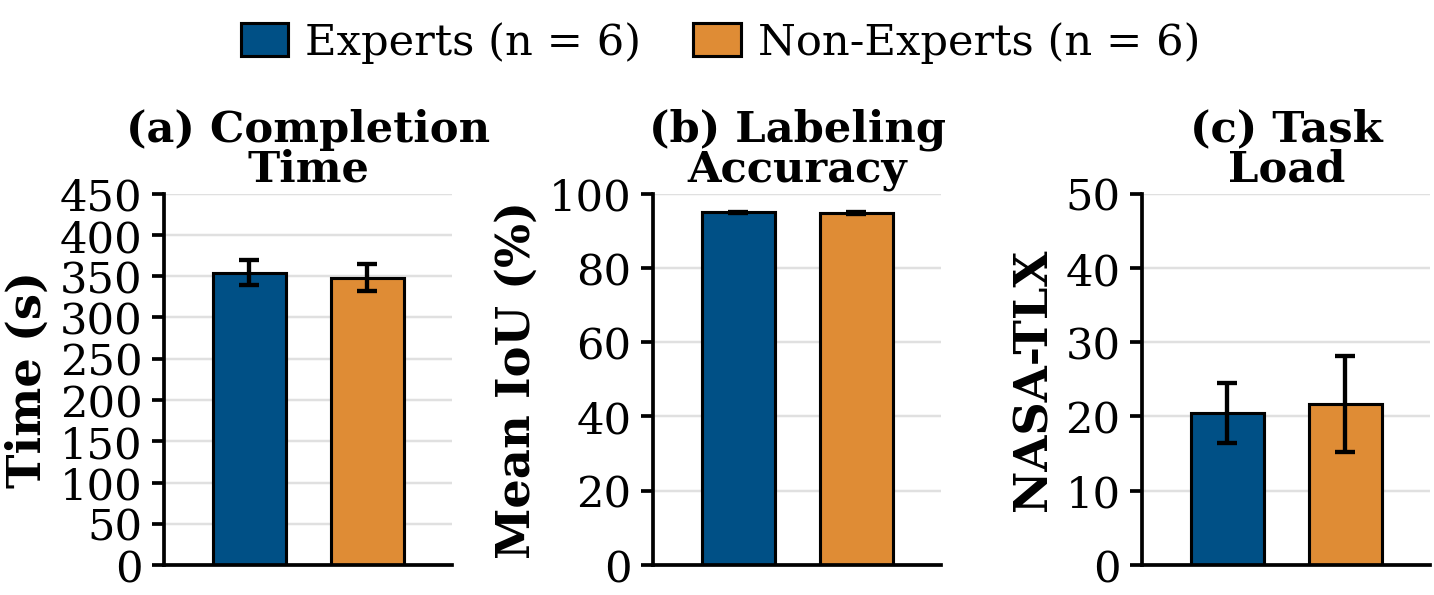}
\caption{User study results by expertise ($n{=}6$ per group). 
Intervals overlap on all three measures.}
\label{fig:user-study-overall}
\vspace{-2mm}
\end{figure}

\begin{figure}[!t]
\centering
\includegraphics[width=\linewidth]{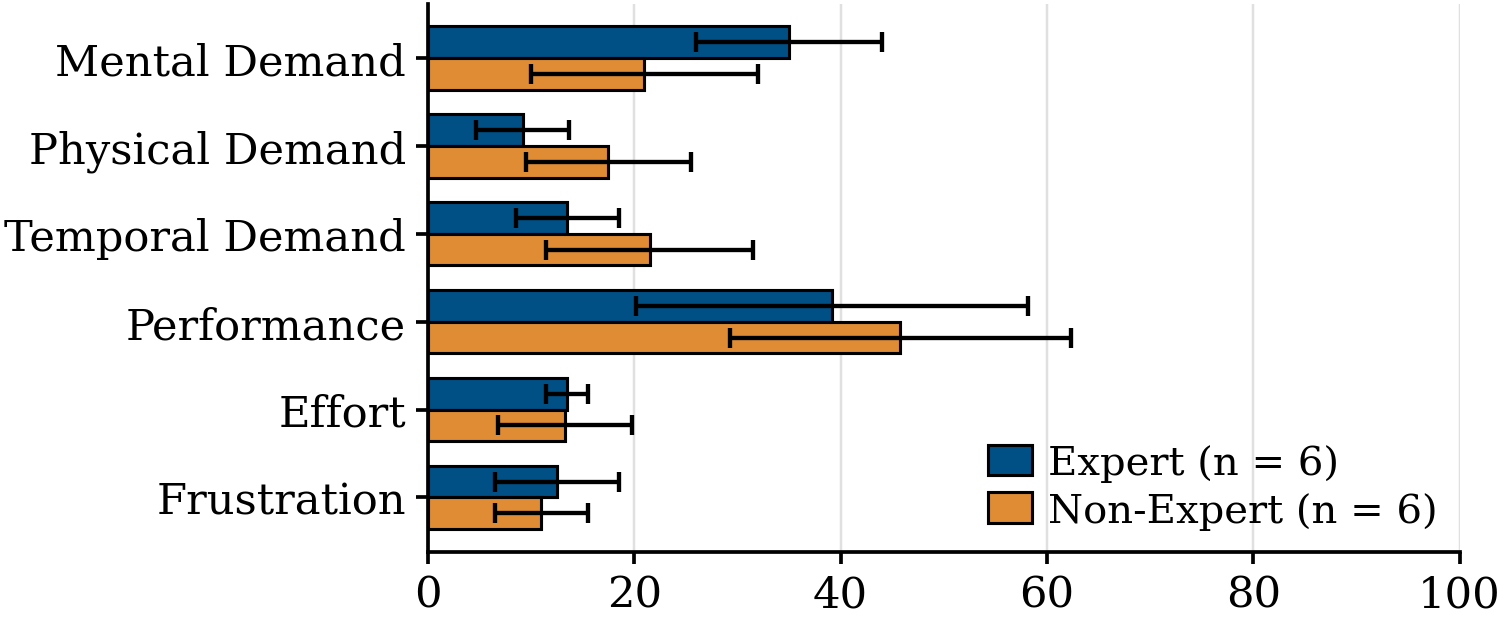}
\caption{NASA-TLX by subscale. The two profiles overlap on every 
subscale, so the agreement in 
Fig.~\ref{fig:user-study-overall}(c) is not an averaging artifact.}
\label{fig:user-study-tlx}
\vspace{-5mm}
\end{figure}

\textbf{Results.} Participants reached a mean box IoU of 94.87\% 
(SD~0.49) at a mean NASA-TLX of 21.04 out of 100 (SD~12.56), 
accurate prompting at low reported task load. HumanPlay itself takes 
14.17\,s (SD~3.15), above the 5--10\,s window in 
Table~\ref{tbl:timing} as it includes setup. Experts and non-experts 
performed comparably on all three measures 
(Fig.~\ref{fig:user-study-overall}), and their subscale profiles 
overlap (Fig.~\ref{fig:user-study-tlx}): intervals overlap on every 
dimension. Annotation effort is near the floor: participants needed a 
median of one gaze point prompt per object, 1.03 on average and never 
more than three.

These results validate a core system requirement: gaze-and-voice 
annotation in MR is accurate and low-burden without robotics 
expertise. The low NASA-TLX scores (21 out of 100) suggest the 
interface could sustain extended teaching sessions without fatigue, 
which is important when deploying to sites with many failure modes.
Point prompts are converted to bounding boxes before 
propagation (Sec.~\ref{subsec:human-interaction-and-annotation}), so 
minor gaze variance is absorbed and does not affect mask quality.

\section{Limitations and Future Work}
\label{sec:limitations}

The loop assumes a human is available and co-located; it cannot 
self-correct. Automating failure detection (e.g., via prediction 
uncertainty) would reduce human burden but not eliminate it: the 
human provides the supervision signal, and that is a feature, not a 
limitation, when the alternative is silent deployment failures.
Model selection in our experiments uses held-out test data (oracle); 
in true deployment the human judges visually via MR, and formalizing 
this visual selection criterion is an open problem.
Fine-tuning time (10--15\,min) is model-dependent, 
not framework-specific; lighter backbones or parameter-efficient 
methods such as LoRA could reduce this significantly.

\section{Conclusion}

Perception failures in new environments are inevitable; what matters
is how fast the robot recovers. {\coolname} closes this loop in
minutes: a co-located human sees failures through mixed reality,
collects targeted data via HumanPlay and gaze/voice annotation, and
SAM2 converts each ${\sim}$20\,s interaction into hundreds of labeled
frames. Forty-five such interactions triple segmentation accuracy,
improve downstream manipulation, and cause no forgetting, with an
interface that any user can operate without training.
Requiring no cloud, no workstation, and no expert, it is immediately
deployable to a new site, where the robot begins working and improves
as failures arise.
Recent
work treats failures as a signal for correcting robot
\emph{actions}~\cite{liu2023reflect, aha2024}; we show they are
equally valuable for correcting robot \emph{perception}. The same
loop applies to any perception task a human can evaluate visually
(object detection, pose estimation, affordance prediction), and the
MR modalities we exploit (gaze, voice, hand tracking, spatial
anchors) open a path toward teaching manipulation skills themselves.
We will publicly release all data, code~\cite{uois_toolkit}, and the MR
application to support future work in this direction.

\ifanonymous\else
  \section*{Acknowledgement}
This work was supported by the DARPA Perceptually-enabled Task Guidance 
(PTG) Program under contract number HR00112220005, the Sony Research Award 
Program, and the National Science Foundation (NSF) under Grant 
No.~2346528. We thank Sai Haneesh Allu for assistance with the real-world 
experiments.

\fi

\bibliographystyle{IEEEtran}
\bibliography{root}

\begin{thebibliography}{10}
\providecommand{\url}[1]{#1}
\csname url@samestyle\endcsname
\providecommand{\newblock}{\relax}
\providecommand{\bibinfo}[2]{#2}
\providecommand{\BIBentrySTDinterwordspacing}{\spaceskip=0pt\relax}
\providecommand{\BIBentryALTinterwordstretchfactor}{4}
\providecommand{\BIBentryALTinterwordspacing}{\spaceskip=\fontdimen2\font plus
\BIBentryALTinterwordstretchfactor\fontdimen3\font minus
  \fontdimen4\font\relax}
\providecommand{\BIBforeignlanguage}[2]{{%
\expandafter\ifx\csname l@#1\endcsname\relax
\typeout{** WARNING: IEEEtran.bst: No hyphenation pattern has been}%
\typeout{** loaded for the language `#1'. Using the pattern for}%
\typeout{** the default language instead.}%
\else
\language=\csname l@#1\endcsname
\fi
#2}}
\providecommand{\BIBdecl}{\relax}
\BIBdecl

\bibitem{khargonkar2024scenereplica}
N.~Khargonkar, S.~H. Allu, Y.~Lu, J.~J. P, B.~Prabhakaran, Y.~Xiang
  \emph{et~al.}, ``{SceneReplica}: Benchmarking real-world robot manipulation
  by creating replicable scenes,'' in \emph{Proc. IEEE Int. Conf. Robot. Autom.
  (ICRA)}.\hskip 1em plus 0.5em minus 0.4em\relax IEEE, 2024, pp. 8258--8264.

\bibitem{liu2024okrobot}
P.~Liu, Y.~Orru, J.~Vakil, C.~Paxton, N.~M.~M. Shafiullah, and L.~Pinto,
  ``{OK-Robot}: What really matters in integrating open-knowledge models for
  robotics,'' in \emph{Robotics: Science and Systems (RSS)}, 2024.

\bibitem{liu2023reflect}
Z.~Liu, A.~Bahety, and S.~Song, ``{REFLECT}: Summarizing robot experiences for
  failure explanation and correction,'' in \emph{Proc. Conf. Robot Learn.
  (CoRL)}, 2023.

\bibitem{aha2024}
J.~Duan, W.~Pumacay, N.~Kumar, Y.~R. Wang, S.~Tian, W.~Yuan, R.~Krishna,
  D.~Fox, A.~Mandlekar, and Y.~Guo, ``{AHA}: A vision-language-model for
  detecting and reasoning over failures in robotic manipulation,'' \emph{arXiv
  preprint arXiv:2410.00371}, 2024.

\bibitem{zeng2026vifailback}
X.~Zeng, X.~Zhou, Y.~Li, J.~Shi, T.~Li, L.~Chen, L.~Ren, and Y.-L. Li,
  ``Diagnose, correct, and learn from manipulation failures via visual
  symbols,'' in \emph{Proc. IEEE/CVF Conf. Comput. Vis. Pattern Recognit.
  (CVPR)}, 2026.

\bibitem{sundermeyer2021cgnet}
M.~Sundermeyer, A.~Mousavian, R.~Triebel, and D.~Fox, ``{Contact-GraspNet}:
  Efficient 6-dof grasp generation in cluttered scenes,'' in \emph{Proc. IEEE
  Int. Conf. Robot. Autom. (ICRA)}.\hskip 1em plus 0.5em minus 0.4em\relax
  IEEE, 2021, pp. 13\,438--13\,444.

\bibitem{xiang2024gto}
Y.~Xiang, S.~H. Allu, R.~Peddi, T.~Summers, and V.~Gogate, ``Grasping
  trajectory optimization with point clouds,'' in \emph{Proc. IEEE/RSJ Int.
  Conf. Intell. Robots Syst. (IROS)}.\hskip 1em plus 0.5em minus 0.4em\relax
  IEEE, 2024, pp. 9885--9892.

\bibitem{ravi2024sam2}
N.~Ravi, V.~Gabeur, Y.-T. Hu, R.~Hu, C.~Ryali, T.~Ma, H.~Khedr, R.~R{\"a}dle,
  C.~Rolland, L.~Gustafson \emph{et~al.}, ``{SAM~2}: Segment anything in images
  and videos,'' in \emph{Proc. Int. Conf. Learn. Represent. (ICLR)}, 2025.

\bibitem{XieCoRL19}
\BIBentryALTinterwordspacing
C.~Xie, Y.~Xiang, A.~Mousavian, and D.~Fox, ``The best of both modes:
  Separately leveraging rgb and depth for unseen object instance
  segmentation,'' in \emph{Proc. Conf. Robot Learn. (CoRL)}, 2019. [Online].
  Available: \url{https://arxiv.org/abs/1907.13236}
\BIBentrySTDinterwordspacing

\bibitem{xie2021unseen}
Y.~Xiang, A.~Mousavian, and D.~Fox, ``Unseen object instance segmentation for
  robotic environments,'' \emph{IEEE Trans. Robot.}, vol.~37, no.~5, pp.
  1343--1359, 2021.

\bibitem{msmformer}
Y.~Lu, Y.~Chen, N.~Ruozzi, and Y.~Xiang, ``Mean shift mask transformer for
  unseen object instance segmentation,'' in \emph{Proc. IEEE Int. Conf. Robot.
  Autom. (ICRA)}.\hskip 1em plus 0.5em minus 0.4em\relax IEEE, 2024, pp.
  2760--2766.

\bibitem{ross2011reduction}
S.~Ross, G.~J. Gordon, and J.~A. Bagnell, ``A reduction of imitation learning
  and structured prediction to no-regret online learning,'' in
  \emph{International Conference on Artificial Intelligence and Statistics
  (AISTATS)}, 2011.

\bibitem{kelly2019hgdagger}
M.~Kelly, C.~Sidrane, K.~Driggs-Campbell, and M.~J. Kochenderfer,
  ``{HG-DAgger}: Interactive imitation learning with human experts,'' in
  \emph{Proc. IEEE Int. Conf. Robot. Autom. (ICRA)}, 2019.

\bibitem{hoque2021thriftydagger}
R.~Hoque, A.~Balakrishna, E.~Novoseller, A.~Wilcox, D.~S. Brown, and
  K.~Goldberg, ``{ThriftyDAgger}: Budget-aware novelty and risk gating for
  interactive imitation learning,'' in \emph{Proc. Conf. Robot Learn. (CoRL)},
  2021.

\bibitem{celemin2022interactive}
C.~Celemin, R.~P{\'e}rez-Dattari, E.~Chisari, G.~Franzese, L.~de~Souza~Rosa,
  R.~Prakash, Z.~Ajanovi{\'c}, M.~Ferraz, A.~Valada, and J.~Kober,
  ``Interactive imitation learning in robotics: A survey,'' \emph{Foundations
  and Trends in Robotics}, vol.~10, no. 1-2, 2022.

\bibitem{liu2023sirius}
H.~Liu, S.~Nasiriany, L.~Zhang, Z.~Bao, and Y.~Zhu, ``Robot learning on the
  job: Human-in-the-loop autonomy and learning during deployment,'' \emph{Int.
  J. Robot. Res.}, 2025.

\bibitem{liu2024olaf}
H.~Liu, A.~Chen, Y.~Zhu, A.~Swaminathan, A.~Kolobov, and C.-A. Cheng,
  ``Interactive robot learning from verbal correction,'' in \emph{Robotics:
  Science and Systems (RSS)}, 2024.

\bibitem{luo2024hilserl}
J.~Luo, C.~Xu, J.~Wu, and S.~Levine, ``Precise and dexterous robotic
  manipulation via human-in-the-loop reinforcement learning,'' \emph{arXiv
  preprint arXiv:2410.21845}, 2024.

\bibitem{allu2025modularroboticautonomousexploration}
\BIBentryALTinterwordspacing
S.~H. Allu, I.~Kadosh, T.~Summers, and Y.~Xiang, ``Build once, monitor
  continuously: Persistent semantic mapping via autonomous exploration and
  open-vocabulary object updates,'' 2025. [Online]. Available:
  \url{https://arxiv.org/abs/2409.15493}
\BIBentrySTDinterwordspacing

\bibitem{lu2024nidsnet}
Y.~Lu, J.~J. P, Y.~Guo, N.~Ruozzi, Y.~Xiang \emph{et~al.}, ``Adapting
  pre-trained vision models for novel instance detection and segmentation,'' in
  \emph{Proc. IEEE/RSJ Int. Conf. Intell. Robots Syst. (IROS)}.\hskip 1em plus
  0.5em minus 0.4em\relax IEEE, 2025, pp. 13\,341--13\,348.

\bibitem{bohg2017interactive}
J.~Bohg, K.~Hausman, B.~Sankaran, O.~Brock, D.~Kragic, S.~Schaal, and
  G.~Sukhatme, ``Interactive perception: Leveraging action in perception and
  perception in action,'' \emph{IEEE Trans. Robot.}, vol.~33, no.~6, pp.
  1273--1291, 2017.

\bibitem{lu2023sss}
Y.~Lu, N.~Khargonkar, Z.~Xu, C.~Averill, K.~Palanisamy, K.~Hang, Y.~Guo,
  N.~Ruozzi, and Y.~Xiang, ``Self-supervised unseen object instance
  segmentation via long-term robot interaction,'' in \emph{Robotics: Science
  and Systems (RSS)}, 2023.

\bibitem{zhao2023aloha}
T.~Z. Zhao, V.~Kumar, S.~Levine, and C.~Finn, ``Learning fine-grained bimanual
  manipulation with low-cost hardware,'' in \emph{Robotics: Science and Systems
  (RSS)}, 2023.

\bibitem{mandlekar2019roboturk}
A.~Mandlekar, Y.~Zhu, A.~Garg, J.~Booher, M.~Spero, A.~Tung, J.~Gao, J.~Emmons,
  A.~Gupta, E.~Orbay, S.~Savarese, and L.~Fei-Fei, ``{RoboTurk}: A
  crowdsourcing platform for robotic skill learning through imitation,'' in
  \emph{Proc. Conf. Robot Learn. (CoRL)}, 2019.

\bibitem{kareer2024egomimic}
S.~Kareer, D.~Patel, R.~Punamiya, P.~Mathur, S.~Cheng, C.~Wang, J.~Hoffman, and
  D.~Xu, ``{EgoMimic}: Scaling imitation learning via egocentric video,'' in
  \emph{Proc. Conf. Robot Learn. (CoRL)}, 2024.

\bibitem{chi2024umi}
C.~Chi, Z.~Xu, C.~Pan, E.~Cousineau, B.~Burchfiel, S.~Feng, R.~Tedrake, and
  S.~Song, ``Universal manipulation interface: In-the-wild robot teaching
  without in-the-wild robots,'' in \emph{Robotics: Science and Systems (RSS)},
  2024.

\bibitem{rise2024}
M.~Kimhi, D.~Vainshtein, C.~Baskin, and D.~Di~Castro, ``Robot instance
  segmentation with few annotations for grasping,'' \emph{arXiv preprint
  arXiv:2407.01302}, 2024.

\bibitem{zeng2020transporter}
A.~Zeng, P.~Florence, J.~Tompson, S.~Welker, J.~Chien, M.~Attarian,
  T.~Armstrong, I.~Krasin, D.~Duber, V.~Sindhwani, and J.~Lee, ``Transporter
  networks: Rearranging the visual world for robotic manipulation,'' in
  \emph{Proc. Conf. Robot Learn. (CoRL)}, 2020.

\bibitem{padalunkal2024protoclip}
J.~J. P, K.~Palanisamy, Y.-W. Chao, X.~Du, and Y.~Xiang, ``{Proto-CLIP:
  Vision-Language Prototypical Network for Few-Shot Learning},'' in \emph{Proc.
  IEEE/RSJ Int. Conf. Intell. Robots Syst. (IROS)}, 2024, pp. 2594--2601.

\bibitem{khazatsky2024droid}
A.~Khazatsky, K.~Pertsch, S.~Nair, A.~Balakrishna, S.~Dasari, S.~Karamcheti,
  S.~Nasiriany \emph{et~al.}, ``{DROID}: A large-scale in-the-wild robot
  manipulation dataset,'' in \emph{Robotics: Science and Systems (RSS)}, 2024.

\bibitem{openxembodiment2024}
{Open X-Embodiment Collaboration}, ``{Open X-Embodiment}: Robotic learning
  datasets and rt-x models,'' in \emph{Proc. IEEE Int. Conf. Robot. Autom.
  (ICRA)}, 2024.

\bibitem{tobin2017domain}
J.~Tobin, R.~Fong, A.~Ray, J.~Schneider, W.~Zaremba, and P.~Abbeel, ``Domain
  randomization for transferring deep neural networks from simulation to the
  real world,'' in \emph{Proc. IEEE/RSJ Int. Conf. Intell. Robots Syst.
  (IROS)}, 2017.

\bibitem{schreiber2025doyouknow}
A.~Schreiber and K.~Driggs-Campbell, ``Do you know the way? human-in-the-loop
  understanding for fast traversability estimation in mobile robotics,''
  \emph{IEEE Robot. Autom. Lett.}, 2025.

\bibitem{rosen2020communicating}
E.~Rosen, D.~Whitney, E.~Phillips, G.~Chien, J.~Tompkin, G.~Konidaris, and
  S.~Tellex, ``Communicating robot arm motion intent through mixed reality
  head-mounted displays,'' in \emph{Robotics Research: The 18th International
  Symposium ISRR}.\hskip 1em plus 0.5em minus 0.4em\relax Springer, 2020, pp.
  301--316.

\bibitem{arunachalam2023holo}
S.~P. Arunachalam, I.~G{\"u}zey, S.~Chintala, and L.~Pinto, ``{Holo-Dex}:
  Teaching dexterity with immersive mixed reality,'' in \emph{Proc. IEEE Int.
  Conf. Robot. Autom. (ICRA)}.\hskip 1em plus 0.5em minus 0.4em\relax IEEE,
  2023, pp. 5962--5969.

\bibitem{liu2025egozero}
V.~Liu, A.~Adeniji, H.~Zhan, S.~Haldar, R.~Bhirangi, P.~Abbeel, and L.~Pinto,
  ``{EgoZero}: Robot learning from smart glasses,'' \emph{arXiv preprint
  arXiv:2505.20290}, 2025.

\bibitem{bohus2022continual}
D.~Bohus, S.~Andrist, A.~Feniello, N.~Saw, and E.~Horvitz, ``Continual learning
  about objects in the wild: An interactive approach,'' in \emph{International
  Conference on Multimodal Interaction (ICMI)}, 2022, pp. 73--82.

\bibitem{padalunkal2023fewsol}
J.~J. P, Y.-W. Chao, and Y.~Xiang, ``{FewSOL: A Dataset for Few-Shot Object
  Learning in Robotic Environments},'' in \emph{Proc. IEEE Int. Conf. Robot.
  Autom. (ICRA)}, 2023, pp. 9140--9146.

\bibitem{ocid}
M.~Suchi, T.~Patten, D.~Fischinger, and M.~Vincze, ``{EasyLabel}: A
  semi-automatic pixel-wise object annotation tool for creating robotic rgb-d
  datasets,'' in \emph{Proc. IEEE Int. Conf. Robot. Autom. (ICRA)}.\hskip 1em
  plus 0.5em minus 0.4em\relax IEEE, 2019, pp. 6678--6684.

\bibitem{osd}
A.~Richtsfeld, T.~M{\"o}rwald, J.~Prankl, M.~Zillich, and M.~Vincze,
  ``Segmentation of unknown objects in indoor environments,'' in \emph{Proc.
  IEEE/RSJ Int. Conf. Intell. Robots Syst. (IROS)}.\hskip 1em plus 0.5em minus
  0.4em\relax IEEE, 2012, pp. 4791--4796.

\bibitem{qian2024riseg}
H.~H. Qian, Y.~Lu, K.~Ren, G.~Wang, N.~Khargonkar, Y.~Xiang, and K.~Hang,
  ``{RISeg}: Robot interactive object segmentation via body frame-invariant
  features,'' in \emph{Proc. IEEE Int. Conf. Robot. Autom. (ICRA)}.\hskip 1em
  plus 0.5em minus 0.4em\relax IEEE, 2024, pp. 13\,954--13\,960.

\bibitem{qian2025rtriseg}
H.~H. Qian, Y.~Chen, G.~Wang, P.~Chanrungmaneekul, and K.~Hang, ``{rt-RISeg}:
  Real-time model-free robot interactive segmentation for active instance-level
  object understanding,'' in \emph{Proc. IEEE/RSJ Int. Conf. Intell. Robots
  Syst. (IROS)}.\hskip 1em plus 0.5em minus 0.4em\relax IEEE, 2025, pp.
  11\,561--11\,568.

\bibitem{zhang2025zisvfm}
Y.~Zhang, M.~Yin, W.~Bi, H.~Yan, S.~Bian, C.-H. Zhang, and C.~Hua, ``{ZISVFM}:
  Zero-shot object instance segmentation in indoor robotic environments with
  vision foundation models,'' \emph{IEEE Trans. Robot.}, vol.~41, pp.
  1568--1580, 2025.

\bibitem{Liu_gdrnpp}
\BIBentryALTinterwordspacing
X.~Liu, R.~Zhang, C.~Zhang, G.~Wang, J.~Tang, Z.~Li, and X.~Ji, ``{GDRNPP}: A
  geometry-guided and fully learning-based object pose estimator,'' \emph{IEEE
  Trans. Pattern Anal. Mach. Intell.}, vol.~47, no.~7, p. 5742–5759, Jul.
  2025. [Online]. Available: \url{http://dx.doi.org/10.1109/TPAMI.2025.3553485}
\BIBentrySTDinterwordspacing

\bibitem{sucan2012OMPL}
I.~A. Sucan, M.~Moll, and L.~E. Kavraki, ``The open motion planning library,''
  \emph{IEEE Robotics \& Automation Magazine}, vol.~19, no.~4, pp. 72--82,
  2012.

\bibitem{hart1988nasatlx}
S.~G. Hart and L.~E. Staveland, ``Development of {NASA-TLX} (task load index):
  Results of empirical and theoretical research,'' in \emph{Human Mental
  Workload}.\hskip 1em plus 0.5em minus 0.4em\relax North-Holland, 1988, pp.
  139--183.

\bibitem{uois_toolkit}
\BIBentryALTinterwordspacing
J.~J. P, A.~Aggarwal, and A.~Maheshwari, ``{uois\_toolkit}: A {PyTorch} toolkit
  for unseen object instance segmentation,'' 2025. [Online]. Available:
  \url{https://github.com/jishnujayakumar/uois_toolkit}
\BIBentrySTDinterwordspacing

\bibitem{p2024robokit}
J.~J. P, ``Robokit: A toolkit for robotic tasks,'' 2024,
  \url{https://github.com/jishnujayakumar/robokit}.

\bibitem{cao2024uois-sam}
R.~Cao, C.~Song, B.~Yang, J.~Wang, P.-A. Heng, and Y.-H. Liu, ``Adapting
  segment anything model for unseen object instance segmentation,'' \emph{arXiv
  preprint arXiv:2409.15481}, 2024.

\end{thebibliography}

\ifappendix
  \clearpage
  \appendices
  \section*{Supplementary Material}
  \subsection{HumanPlay Data Collection}
\label{app:humanplay}

The data collection pipeline is summarized in
Table~\ref{tab:humanplay_pipeline}. The user performs a HumanPlay
interaction (physically rearranging objects to reduce occlusion) while the
system records a short RGB-D video, then annotates only the final frame
using gaze-based point prompts and voice commands through the MR interface.
SAM2~\cite{ravi2024sam2} converts point prompts to bounding boxes in image
mode and propagates masks to earlier frames in video mode, producing dense
per-frame supervision from a single annotated frame.

\begin{table}[H]
\centering
\renewcommand{\arraystretch}{1.2}
\rowcolors{2}{gray!8}{white}
\resizebox{\linewidth}{!}{
\begin{tabular}{@{}r l@{}}
\toprule
\rowcolor{headerpeach}
\textbf{Step} & \textbf{Operation} \\
\midrule
1. Failure observation & Human identifies incorrect masks in robot prediction \\
2. HumanPlay interaction & Human repositions objects to reduce clutter \\
3. Video capture & Short RGB-D sequence recorded \\
4. Final-frame annotation & Gaze-based point prompts on last frame \\
5. BBox generation & SAM2~\cite{ravi2024sam2} image mode converts points to bounding boxes \\
6. Mask propagation & SAM2~\cite{ravi2024sam2} video mode propagates masks to earlier frames using Robokit~\cite{p2024robokit} \\
7. Dense supervision & Sparse annotation converted to frame-level masks \\
\bottomrule
\end{tabular}}
\caption{{\coolname}-HumanPlay data collection pipeline.}
\label{tab:humanplay_pipeline}
\end{table}

Fig.~\ref{fig:d5} shows five representative scenes from the
{\coolname}-HumanPlay D5
dataset\footnote{\url{https://utdallas.box.com/v/iTeach-HumanPlay-D5}}.
Fig.~\ref{fig:d40} shows all 40 scenes from the larger D40
dataset\footnote{\url{https://utdallas.box.com/v/iTeach-HumanPlay-D40}}.
Five scenes (2, 8, 16, 21, 33) exhibit mask propagation failures,
typically caused by strong occlusion or large inter-frame object motion.

\begin{figure}[H]
    \centering
    \begin{subfigure}{0.31\linewidth}
        \includegraphics[width=\linewidth]{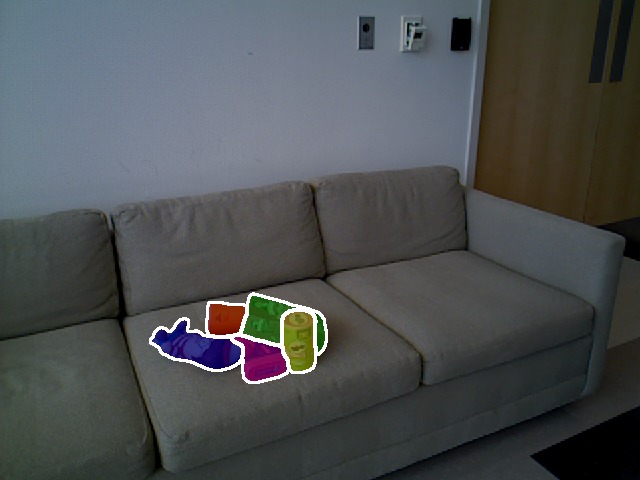}
        \caption*{Scene-1}
    \end{subfigure}\hfill
    \begin{subfigure}{0.31\linewidth}
        \includegraphics[width=\linewidth]{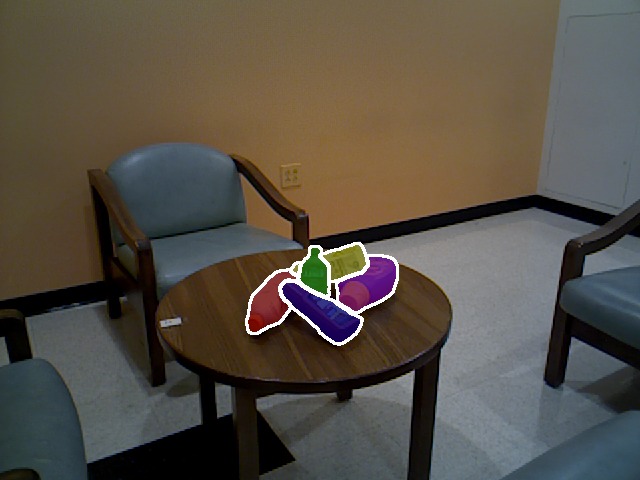}
        \caption*{Scene-2}
    \end{subfigure}\hfill
    \begin{subfigure}{0.31\linewidth}
        \includegraphics[width=\linewidth]{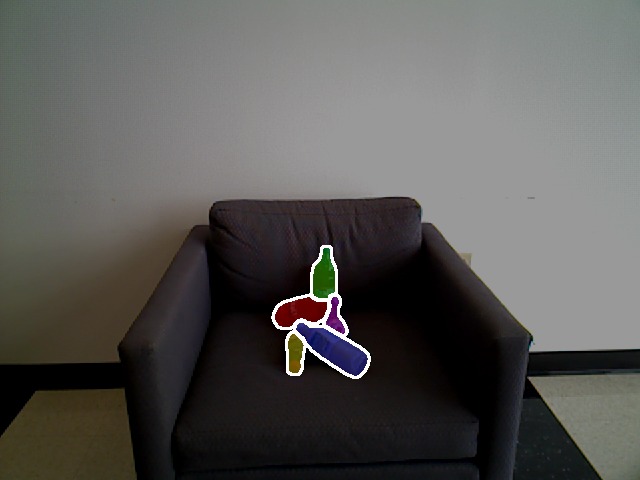}
        \caption*{Scene-3}
    \end{subfigure}\\[6pt]
    \begin{subfigure}{0.31\linewidth}
        \includegraphics[width=\linewidth]{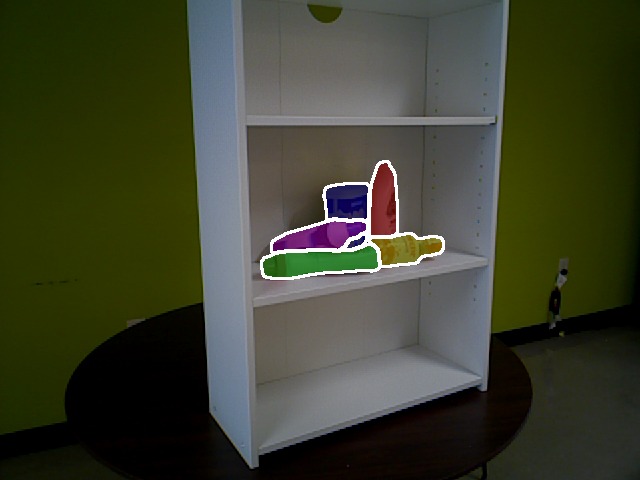}
        \caption*{Scene-4}
    \end{subfigure}\hspace{0.035\linewidth}
    \begin{subfigure}{0.31\linewidth}
        \includegraphics[width=\linewidth]{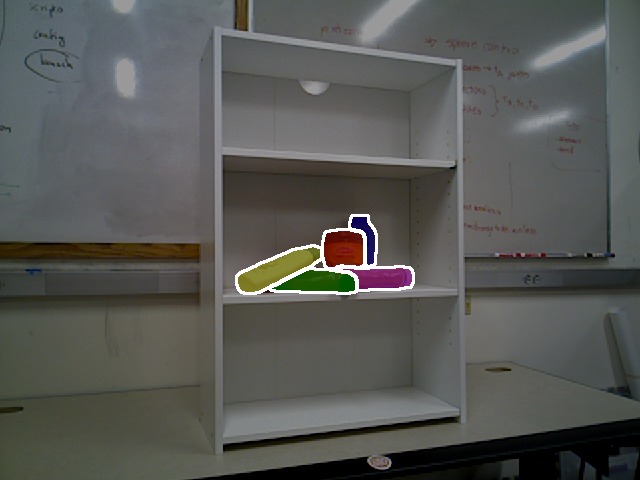}
        \caption*{Scene-5}
    \end{subfigure}
    \vspace{4pt}
    \caption{Five scenes from D5 with ground-truth masks from SAM2~\cite{ravi2024sam2}.}
    \label{fig:d5}
\end{figure}

\begin{figure*}[t]
    \centering
    \begin{subfigure}{0.117\textwidth}\includegraphics[width=\linewidth]{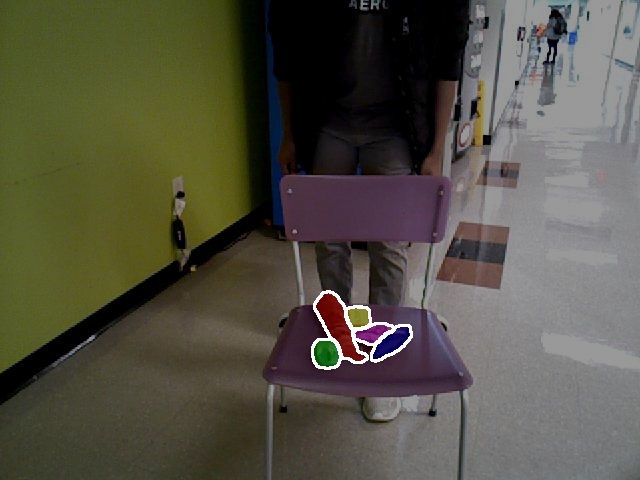}\caption*{1}\end{subfigure}\hfill
    \begin{subfigure}{0.117\textwidth}\includegraphics[width=\linewidth]{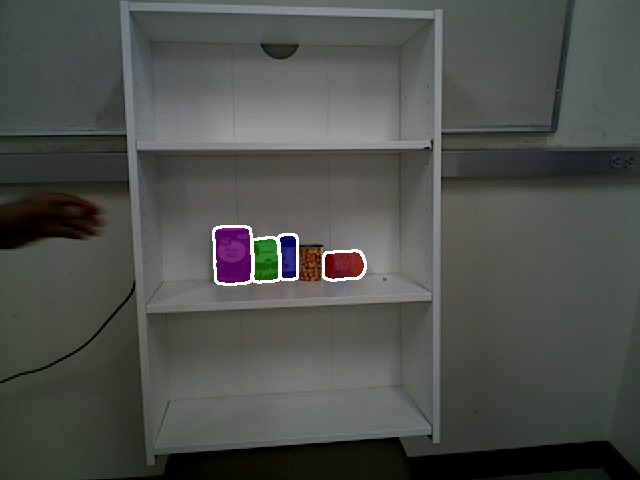}\caption*{\textcolor{red}{\textbf{2}}}\end{subfigure}\hfill
    \begin{subfigure}{0.117\textwidth}\includegraphics[width=\linewidth]{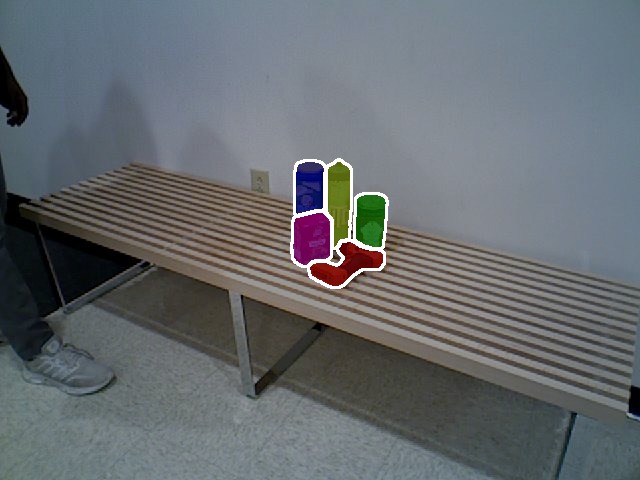}\caption*{3}\end{subfigure}\hfill
    \begin{subfigure}{0.117\textwidth}\includegraphics[width=\linewidth]{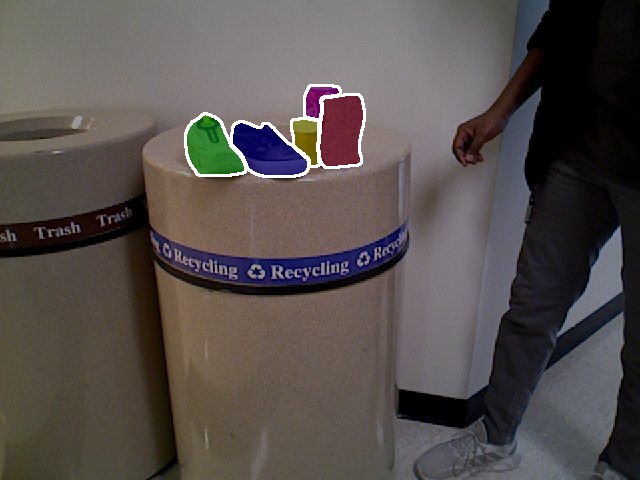}\caption*{4}\end{subfigure}\hfill
    \begin{subfigure}{0.117\textwidth}\includegraphics[width=\linewidth]{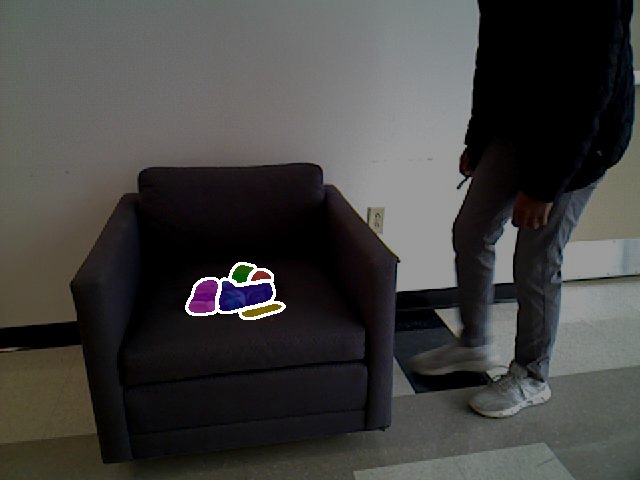}\caption*{5}\end{subfigure}\hfill
    \begin{subfigure}{0.117\textwidth}\includegraphics[width=\linewidth]{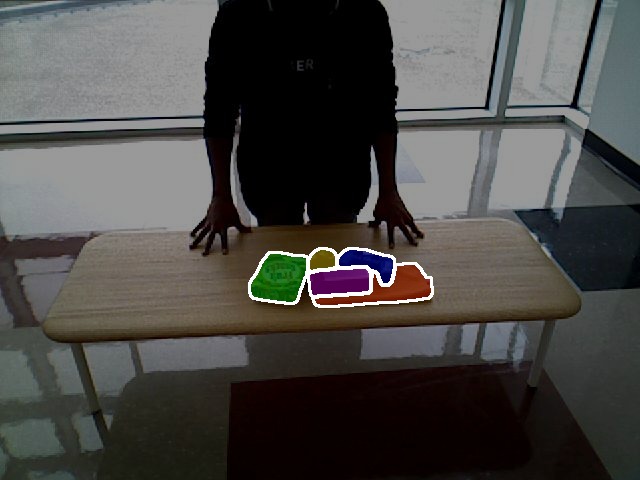}\caption*{6}\end{subfigure}\hfill
    \begin{subfigure}{0.117\textwidth}\includegraphics[width=\linewidth]{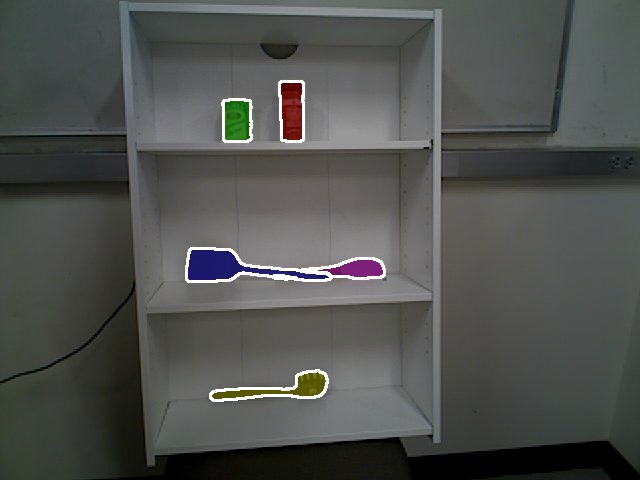}\caption*{7}\end{subfigure}\hfill
    \begin{subfigure}{0.117\textwidth}\includegraphics[width=\linewidth]{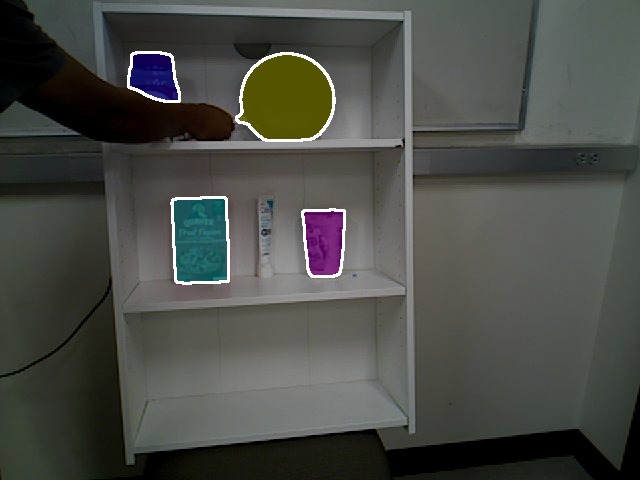}\caption*{\textcolor{red}{\textbf{8}}}\end{subfigure}\\[2pt]
    \begin{subfigure}{0.117\textwidth}\includegraphics[width=\linewidth]{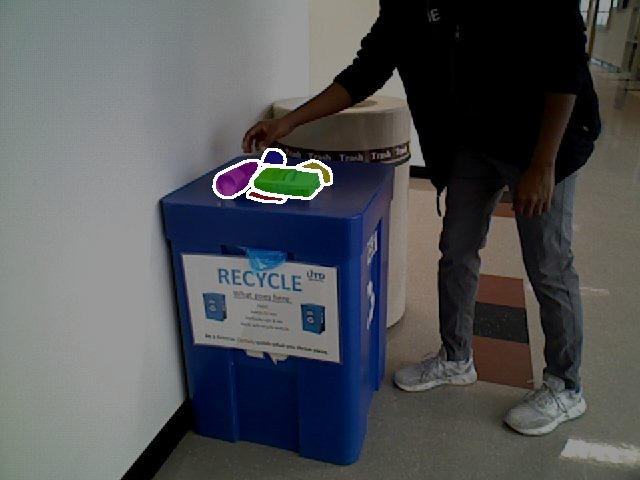}\caption*{9}\end{subfigure}\hfill
    \begin{subfigure}{0.117\textwidth}\includegraphics[width=\linewidth]{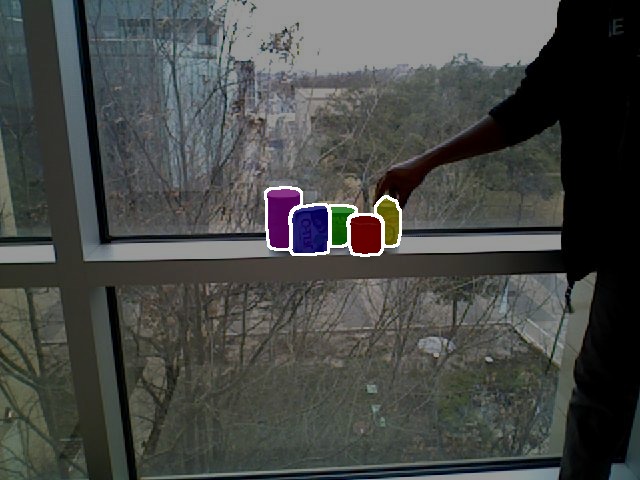}\caption*{10}\end{subfigure}\hfill
    \begin{subfigure}{0.117\textwidth}\includegraphics[width=\linewidth]{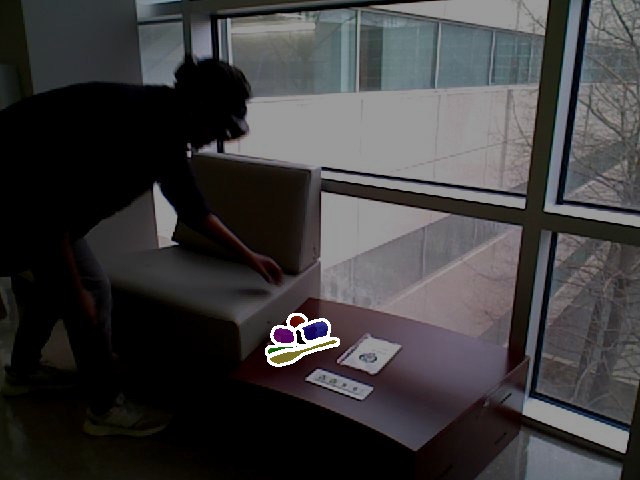}\caption*{11}\end{subfigure}\hfill
    \begin{subfigure}{0.117\textwidth}\includegraphics[width=\linewidth]{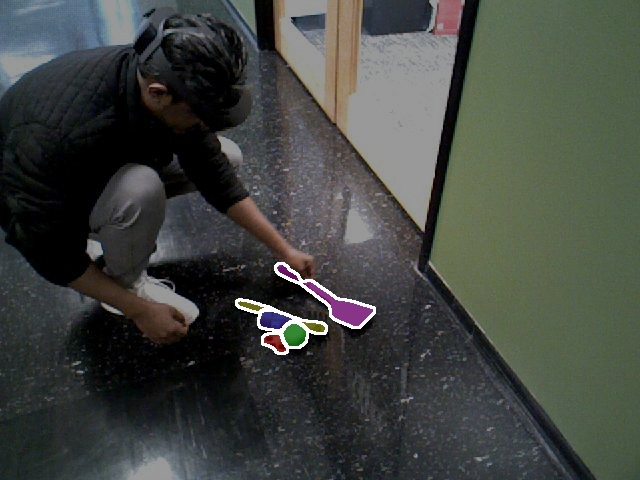}\caption*{12}\end{subfigure}\hfill
    \begin{subfigure}{0.117\textwidth}\includegraphics[width=\linewidth]{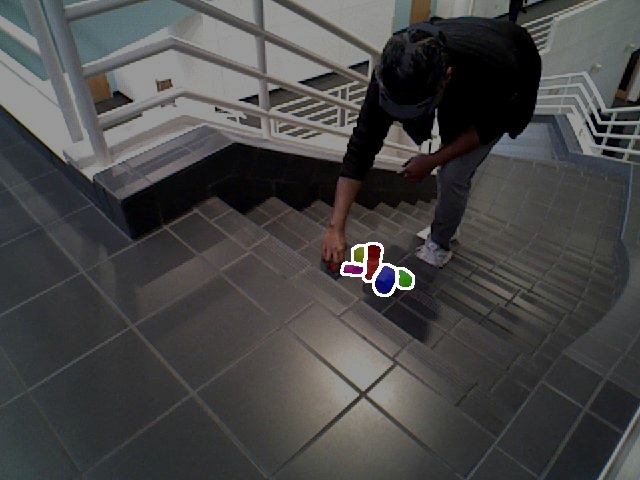}\caption*{13}\end{subfigure}\hfill
    \begin{subfigure}{0.117\textwidth}\includegraphics[width=\linewidth]{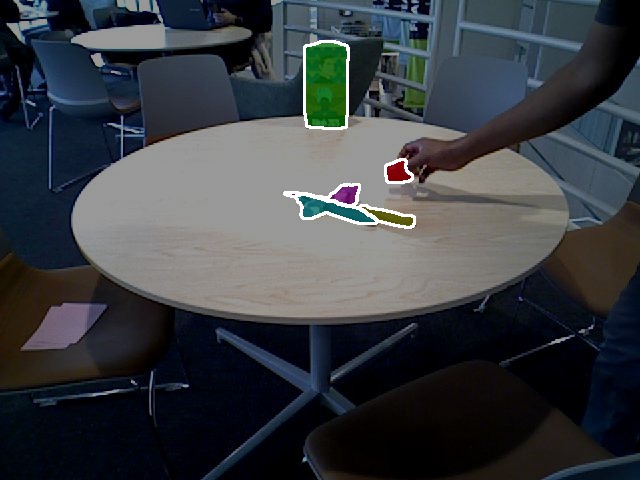}\caption*{14}\end{subfigure}\hfill
    \begin{subfigure}{0.117\textwidth}\includegraphics[width=\linewidth]{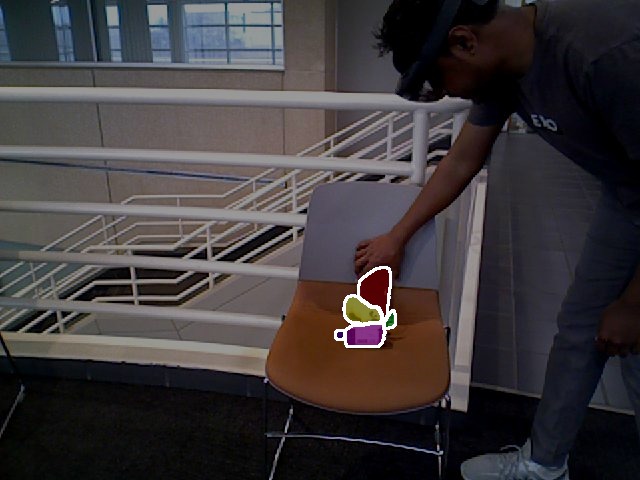}\caption*{15}\end{subfigure}\hfill
    \begin{subfigure}{0.117\textwidth}\includegraphics[width=\linewidth]{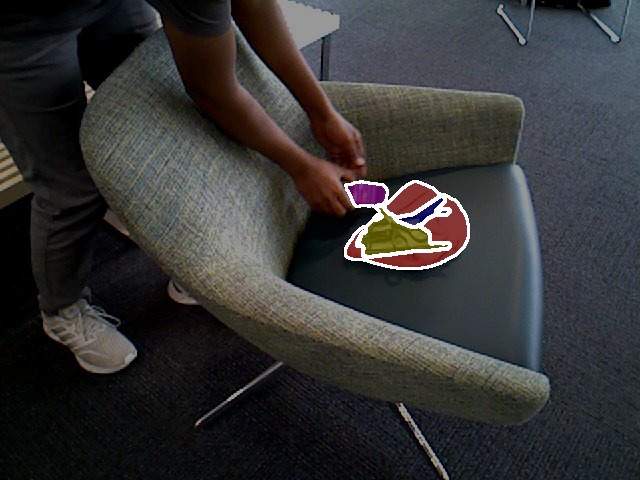}\caption*{\textcolor{red}{\textbf{16}}}\end{subfigure}\\[2pt]
    \begin{subfigure}{0.117\textwidth}\includegraphics[width=\linewidth]{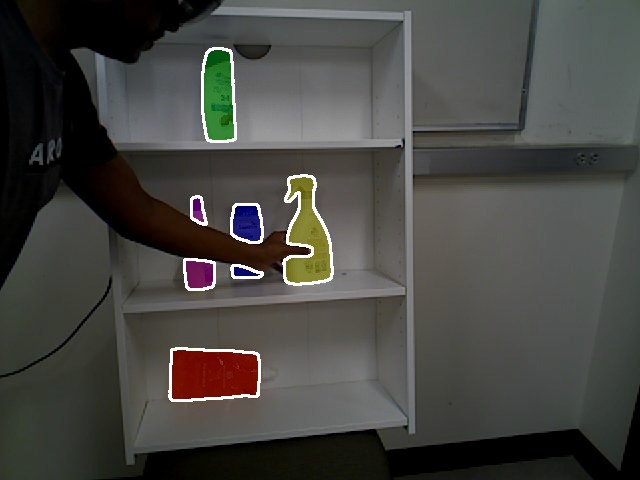}\caption*{17}\end{subfigure}\hfill
    \begin{subfigure}{0.117\textwidth}\includegraphics[width=\linewidth]{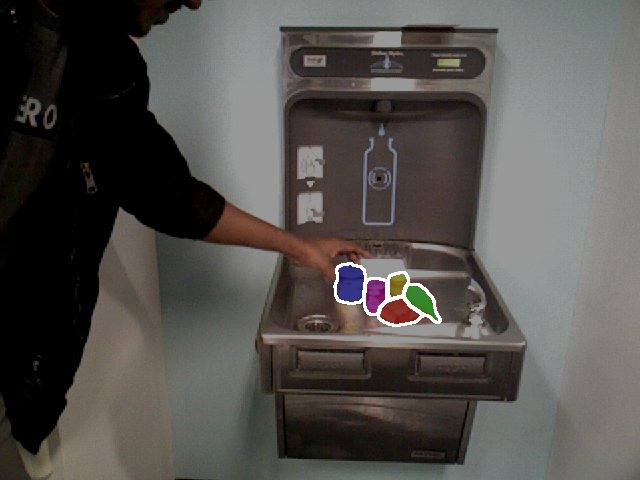}\caption*{18}\end{subfigure}\hfill
    \begin{subfigure}{0.117\textwidth}\includegraphics[width=\linewidth]{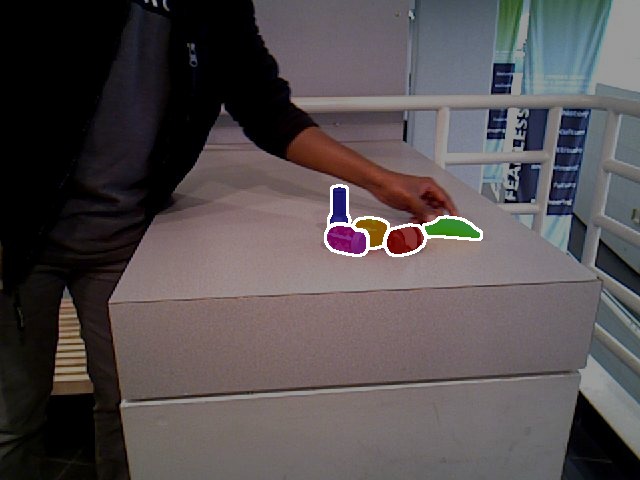}\caption*{19}\end{subfigure}\hfill
    \begin{subfigure}{0.117\textwidth}\includegraphics[width=\linewidth]{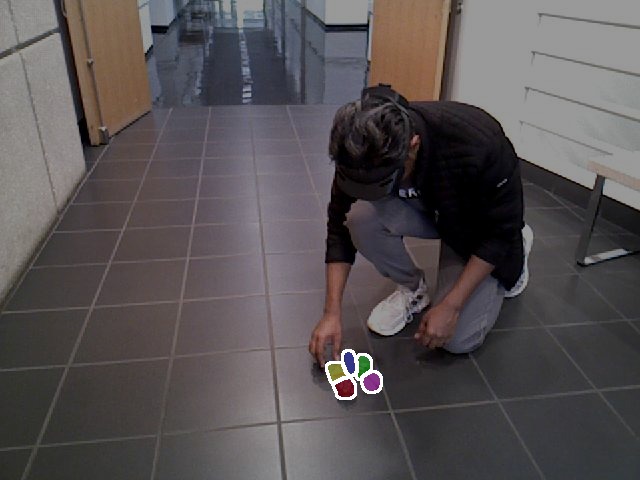}\caption*{20}\end{subfigure}\hfill
    \begin{subfigure}{0.117\textwidth}\includegraphics[width=\linewidth]{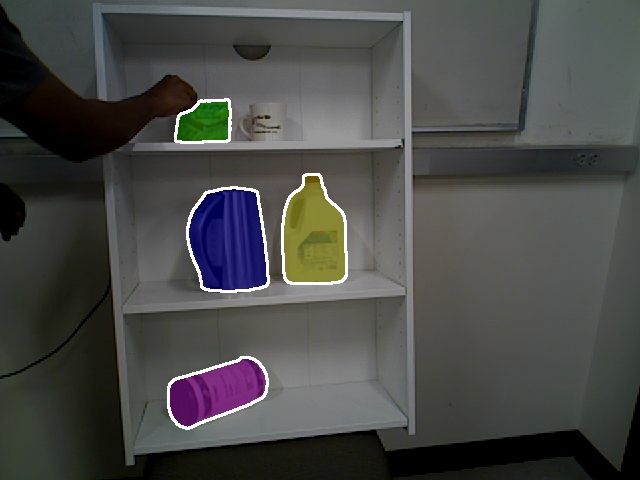}\caption*{\textcolor{red}{\textbf{21}}}\end{subfigure}\hfill
    \begin{subfigure}{0.117\textwidth}\includegraphics[width=\linewidth]{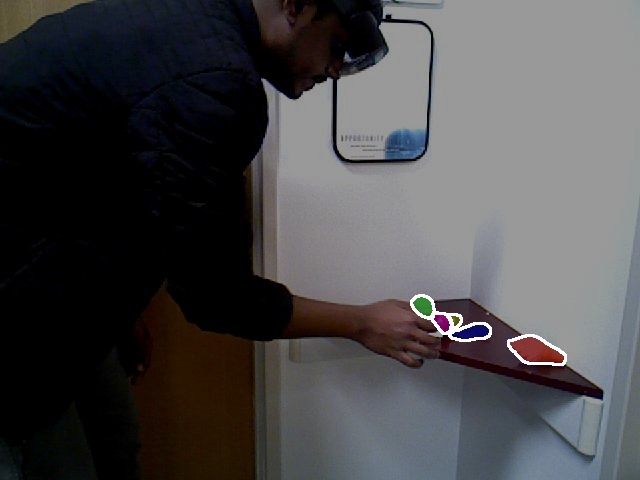}\caption*{22}\end{subfigure}\hfill
    \begin{subfigure}{0.117\textwidth}\includegraphics[width=\linewidth]{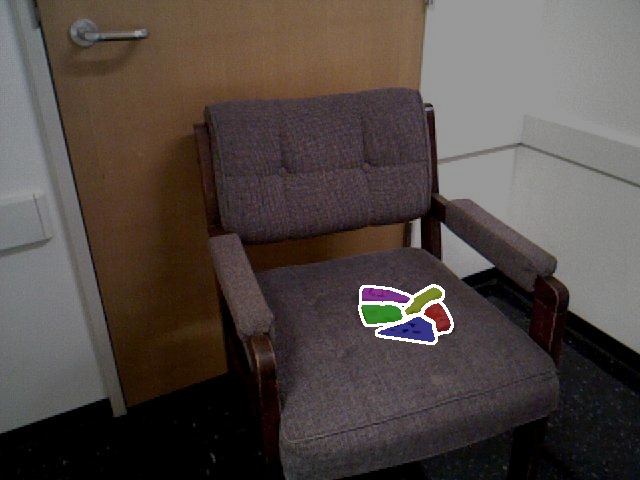}\caption*{23}\end{subfigure}\hfill
    \begin{subfigure}{0.117\textwidth}\includegraphics[width=\linewidth]{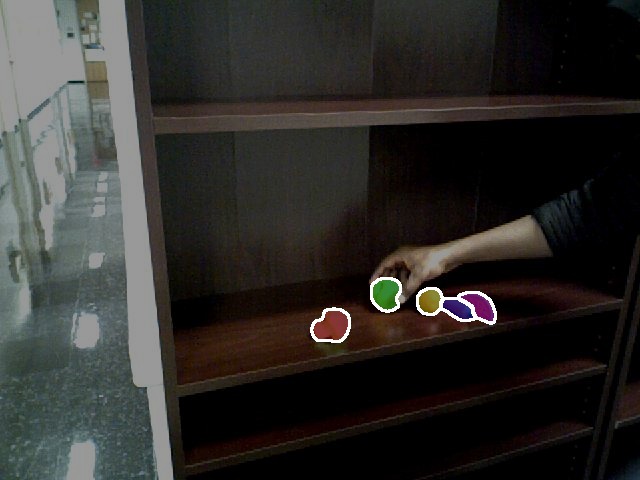}\caption*{24}\end{subfigure}\\[2pt]
    \begin{subfigure}{0.117\textwidth}\includegraphics[width=\linewidth]{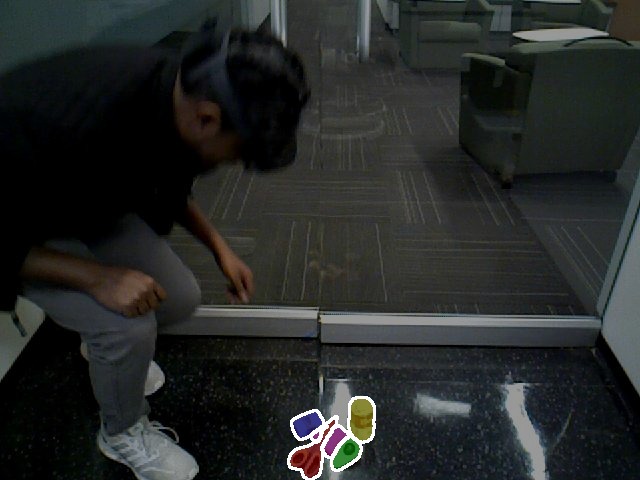}\caption*{25}\end{subfigure}\hfill
    \begin{subfigure}{0.117\textwidth}\includegraphics[width=\linewidth]{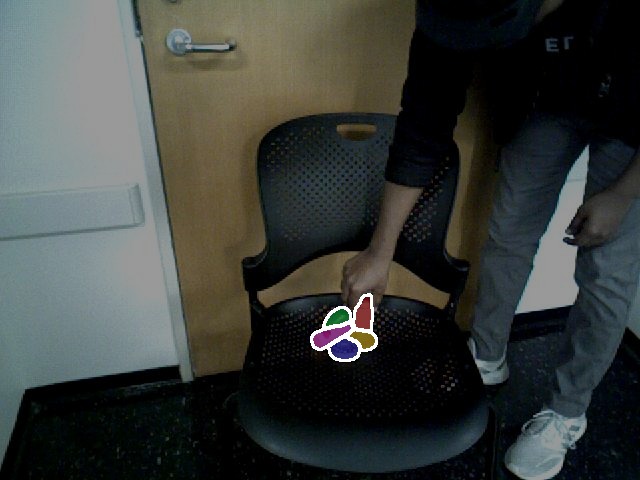}\caption*{26}\end{subfigure}\hfill
    \begin{subfigure}{0.117\textwidth}\includegraphics[width=\linewidth]{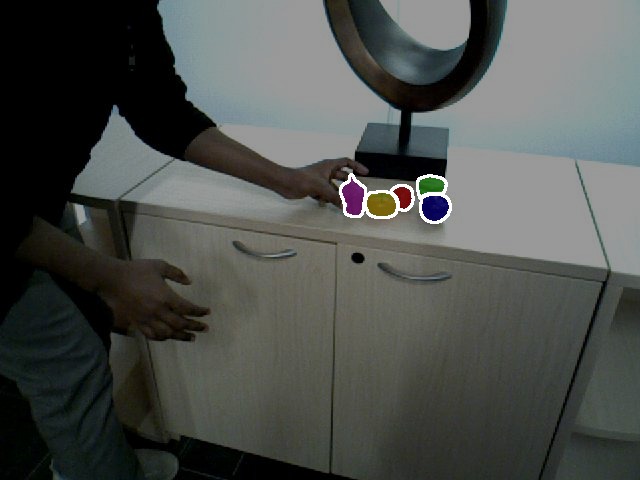}\caption*{27}\end{subfigure}\hfill
    \begin{subfigure}{0.117\textwidth}\includegraphics[width=\linewidth]{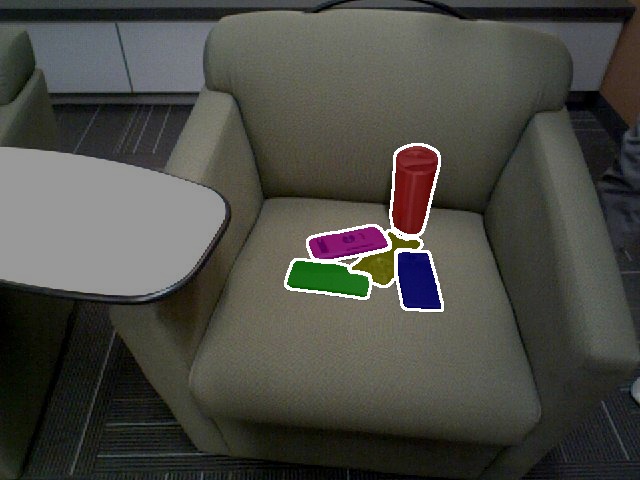}\caption*{28}\end{subfigure}\hfill
    \begin{subfigure}{0.117\textwidth}\includegraphics[width=\linewidth]{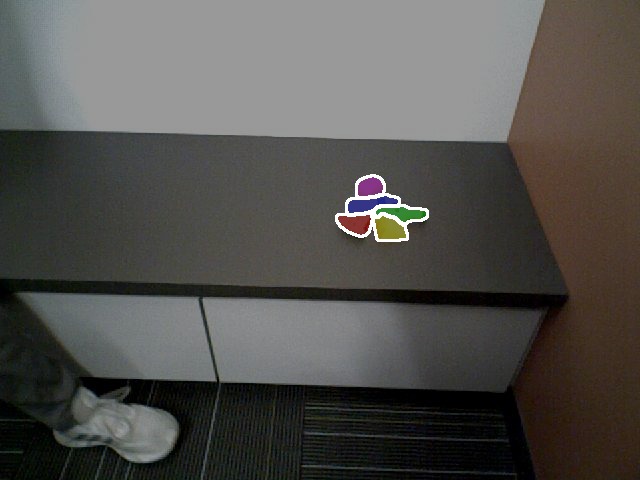}\caption*{29}\end{subfigure}\hfill
    \begin{subfigure}{0.117\textwidth}\includegraphics[width=\linewidth]{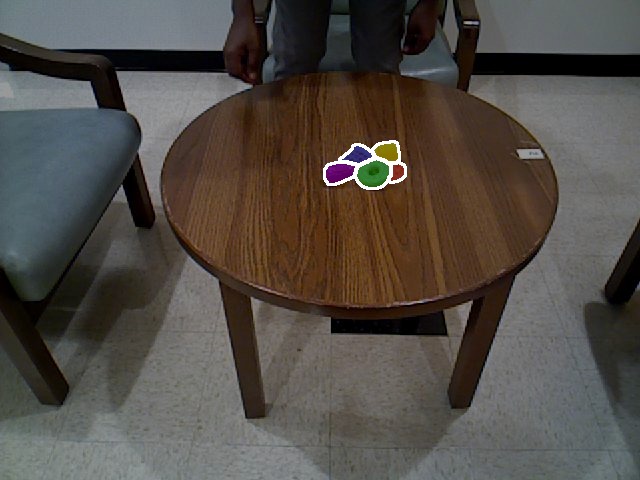}\caption*{30}\end{subfigure}\hfill
    \begin{subfigure}{0.117\textwidth}\includegraphics[width=\linewidth]{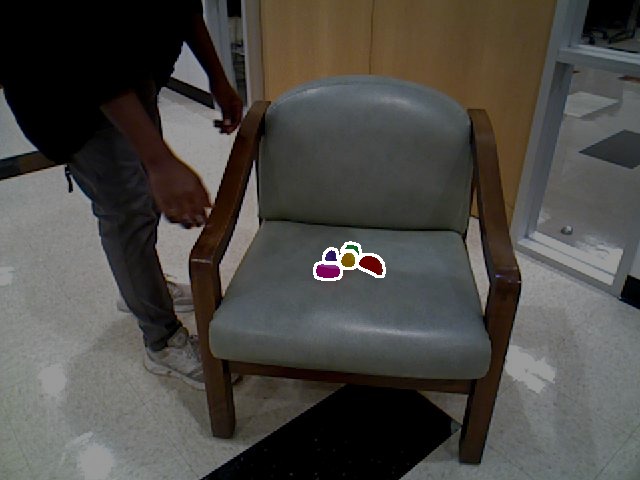}\caption*{31}\end{subfigure}\hfill
    \begin{subfigure}{0.117\textwidth}\includegraphics[width=\linewidth]{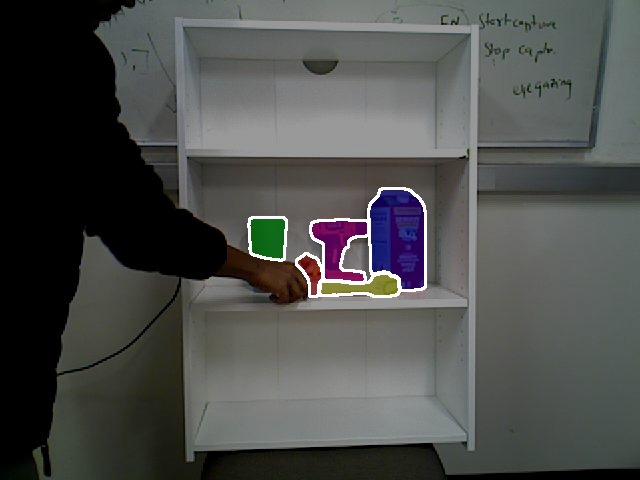}\caption*{32}\end{subfigure}\\[2pt]
    \begin{subfigure}{0.117\textwidth}\includegraphics[width=\linewidth]{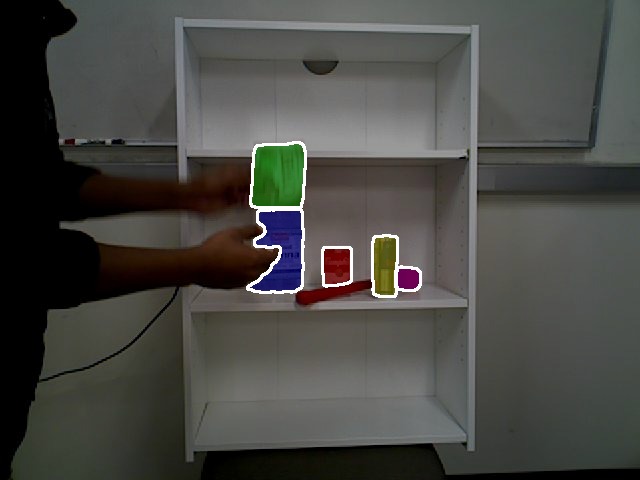}\caption*{\textcolor{red}{\textbf{33}}}\end{subfigure}\hfill
    \begin{subfigure}{0.117\textwidth}\includegraphics[width=\linewidth]{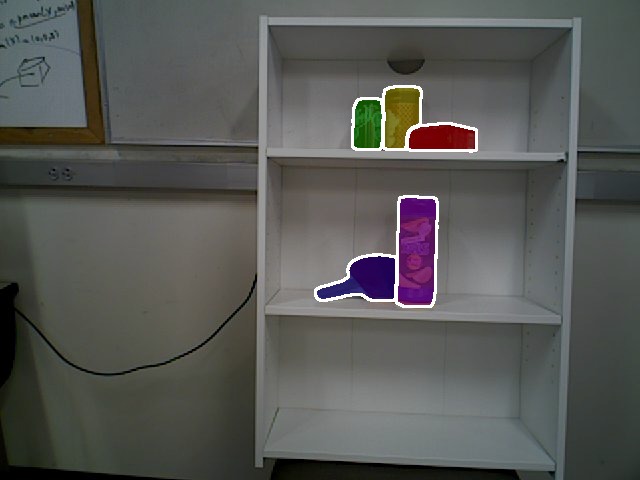}\caption*{34}\end{subfigure}\hfill
    \begin{subfigure}{0.117\textwidth}\includegraphics[width=\linewidth]{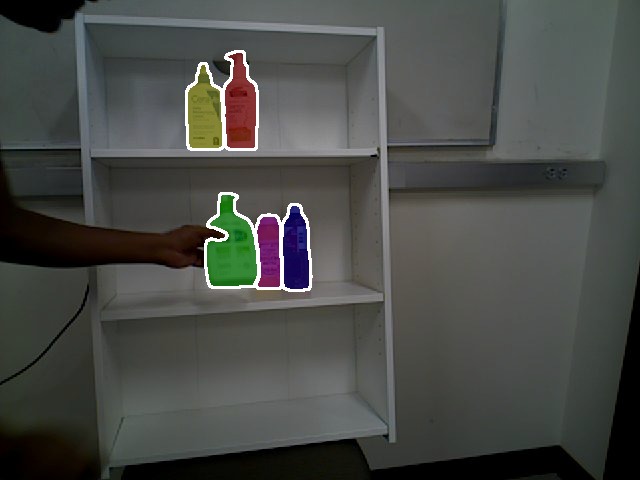}\caption*{35}\end{subfigure}\hfill
    \begin{subfigure}{0.117\textwidth}\includegraphics[width=\linewidth]{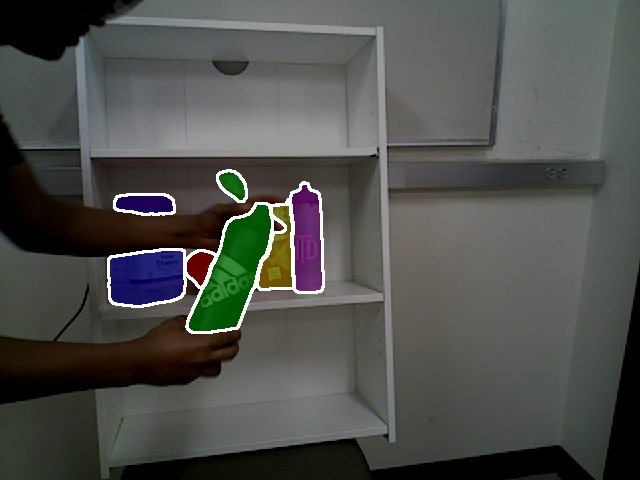}\caption*{36}\end{subfigure}\hfill
    \begin{subfigure}{0.117\textwidth}\includegraphics[width=\linewidth]{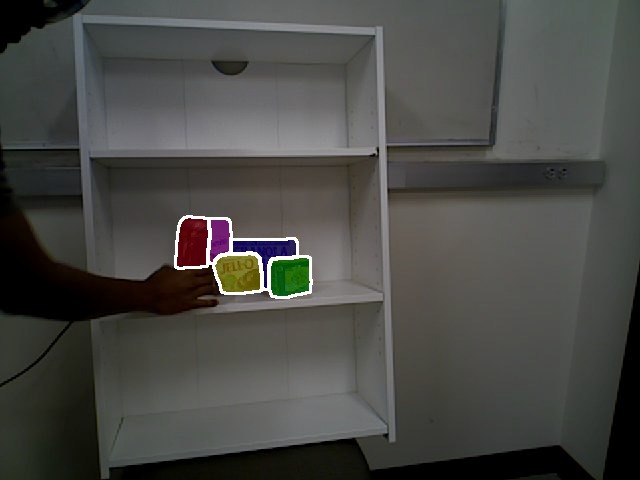}\caption*{37}\end{subfigure}\hfill
    \begin{subfigure}{0.117\textwidth}\includegraphics[width=\linewidth]{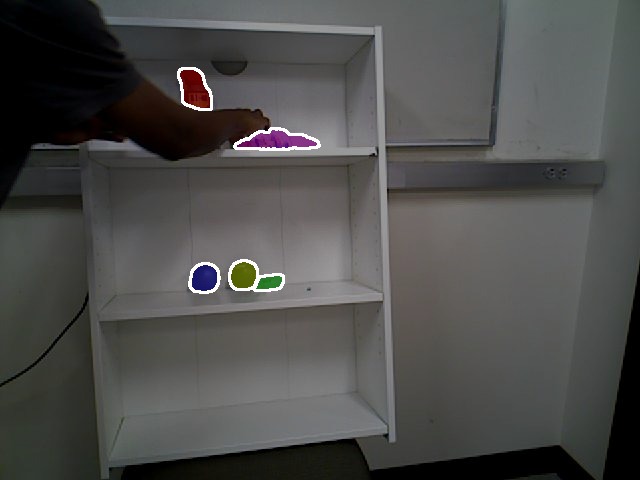}\caption*{38}\end{subfigure}\hfill
    \begin{subfigure}{0.117\textwidth}\includegraphics[width=\linewidth]{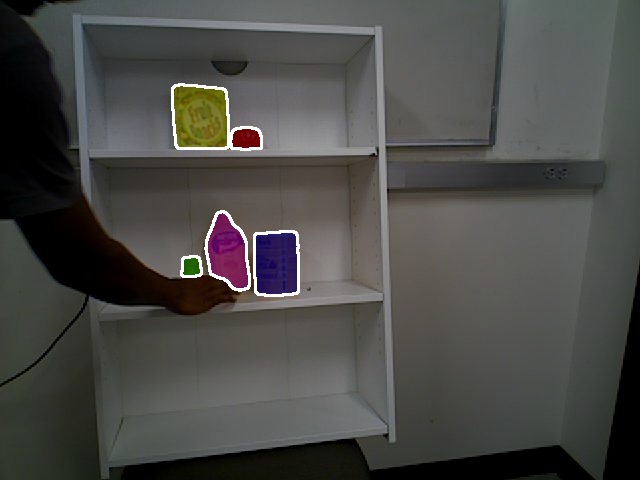}\caption*{39}\end{subfigure}\hfill
    \begin{subfigure}{0.117\textwidth}\includegraphics[width=\linewidth]{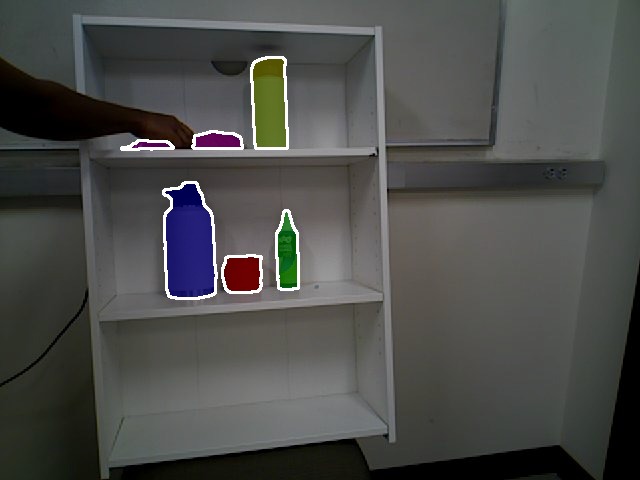}\caption*{40}\end{subfigure}
    \vspace{4pt}
    \caption{All 40 scenes from D40.
    Masks are propagated backward from the annotated final frame.
    Scenes in \textcolor{red}{\textbf{red}} (2, 8, 16, 21, 33) are
    propagation failures due to occlusion or large object motion.}
    \label{fig:d40}
\end{figure*}

  \subsection{Additional Analysis}
\label{app:analysis}

\subsubsection{Data Collection Efficiency}

Table~\ref{tab:data_efficiency} compares data collection strategies.
Robot pushing~\cite{lu2023sss} produces one self-supervised frame per
interaction. Offline labeling requires a full mask per frame. {\coolname}
records a short video, annotates only the final frame, and propagates masks
to all earlier frames.

\begin{table}[H]
\centering
\renewcommand{\arraystretch}{1.2}
\rowcolors{2}{gray!8}{white}
\resizebox{\linewidth}{!}{
\begin{tabular}{@{}lcccc@{}}
\toprule
\rowcolor{headerpeach}
\textbf{Method} & \textbf{Interaction} & \textbf{Data Collected} & \textbf{Annotation Effort} & \textbf{Supervision} \\
\midrule
Robot pushing~\cite{lu2023sss} & Robot motion & Single frame & None & Self-supervised \\
Offline labeling & None & Single frame & Full mask & Dense manual \\
{\coolname} (ours) & 5--10\,s HumanPlay & Video sequence & Last-frame only & FS3 Propagated dense \\
\bottomrule
\end{tabular}}
\caption{Comparison of data collection strategies.}
\label{tab:data_efficiency}
\end{table}

\subsubsection{Choice of Pretrained UOIS Model}

Several UOIS models exist~\cite{cao2024uois-sam, qian2024riseg,
qian2025rtriseg, zhang2025zisvfm, msmformer, xie2021unseen, XieCoRL19}.
We use MSMFormer~\cite{msmformer} for three reasons.
(1)~It has publicly available code and pretrained weights, which is
necessary for deployment-time fine-tuning.
(2)~Lu~et~al.~\cite{lu2023sss} previously fine-tuned the same model with
robot-collected data, providing a direct comparison of data sources: their
autonomous pushing versus our failure-driven collection. Their approach
collects one frame per push and requires the robot to discover
informative interactions autonomously. {\coolname} instead has a
human identify failures, collect short video sequences, and annotate only
the final frame; propagation then yields dense supervision from each
interaction.
(3)~MSMFormer is the perception backbone in SceneReplica~\cite{khargonkar2024scenereplica},
so using the same model lets us measure whether perception gains propagate to
manipulation.

\subsubsection{Human Effort and Adaptation Time}

Each data collection round requires a 5--10\,s HumanPlay interaction
followed by 5--10\,s of gaze-and-voice annotation per object, with 1--3
point prompts per object. The full timing breakdown, split into
system-constant and model-dependent components, is in
Table~\ref{tbl:timing} of the main paper.

\subsubsection{Longer Training / Fine-Tuning}

The main paper uses iterative fine-tuning with 2K-iteration schedules. Here
we aggregate the same HumanPlay data and train for 50K iterations from the
pretrained initialization to isolate the effect of longer training from the
iterative update strategy.

Table~\ref{tab:iteach_stage1_stage2} compares five model variants.
MSMFormer~\cite{msmformer} is the pretrained baseline.
Lu~et~al.~(OCID/OSD)~\cite{lu2023sss} denote MSMFormer fine-tuned on
robot-pushing data from OCID~\cite{ocid} or OSD~\cite{osd} scenes.
{\coolname}-UOIS$^\dagger$ is trained from scratch on HumanPlay data only.
{\coolname}-UOIS$^\ddagger$ is initialized from pretrained MSMFormer and
fine-tuned on the same HumanPlay data.

Following~\cite{msmformer}, let $\hat{M}_i$ be the $i$-th predicted
mask, $M_j$ the $j$-th ground-truth mask, and $\pi(i)$ the Hungarian
matching. We compute overlap precision, recall, and F-measure:
\begin{gather}
P = \frac{\sum_i |\hat{M}_i \cap M_{\pi(i)}|}{\sum_i |\hat{M}_i|},
\qquad
R = \frac{\sum_i |\hat{M}_i \cap M_{\pi(i)}|}{\sum_j |M_j|}\notag\\[4pt]
F = \frac{2PR}{P+R}
\end{gather}
and report object F-measure~($F_o$), boundary F-measure~($F_b$),
and the fraction of objects with $F_o \ge 0.75$~($I_{0.75}$),
combined as:
\begin{equation}
C \;=\; 0.4\,F_o + 0.4\,F_b + 0.2\,I_{0.75}
\end{equation}
$\Delta C$ denotes the change relative to pretrained MSMFormer.

\begingroup
\renewcommand{\arraystretch}{1.4}
\begin{table*}[t]
\centering
\resizebox{\linewidth}{!}{%
\begin{tabular}{@{}c ccccc ccccc ccccc ccccc@{}}
\toprule
\multirow{2}{*}{\textbf{Model}}
& \multicolumn{5}{c}{\textbf{OCID}~\cite{ocid}}
& \multicolumn{5}{c}{\textbf{OSD}~\cite{osd}}
& \multicolumn{5}{c}{\textbf{RobotPushing}~\cite{lu2023sss}}
& \multicolumn{5}{c}{\textbf{\coolname-HumanPlay}} \\
\cmidrule(lr){2-6}\cmidrule(lr){7-11}\cmidrule(lr){12-16}\cmidrule(lr){17-21}
& $F_o$ & $F_b$ & $I_{0.75}$ & $C$ & $\Delta C$
& $F_o$ & $F_b$ & $I_{0.75}$ & $C$ & $\Delta C$
& $F_o$ & $F_b$ & $I_{0.75}$ & $C$ & $\Delta C$
& $F_o$ & $F_b$ & $I_{0.75}$ & $C$ & $\Delta C$ \\
\midrule
\multicolumn{21}{c}{\textit{Stage-1 (Stage-1 fine-tuned)}} \\
\midrule
MSMFormer~\cite{msmformer}
& 88.2 & 82.9 & 83.2 & 84.8 & 0
& 82.6 & 60.9 & 77.1 & 73.5 & 0
& 87.0 & 79.5 & 77.3 & 81.3 & 0
& 28.2 & 21.7 & 30.9 & 26.9 & 0 \\
Lu et al.~\cite{lu2023sss} (OCID) & 89.9 & 85.4 & 82.9 & 86.7 & +1.9 & 84.6 & 69.0 & 76.2 & 76.7 &
+3.2 & 87.8 & 81.2 & 75.1 & 82.6 & +1.3 & 24.3 & 21.9 & 28.4 & 24.1 & -2.8 \\
Lu et al.~\cite{lu2023sss} (OSD)  & 89.8 & 84.5 & 83.0 & 86.3 & +1.5 & 82.1 & 61.6 & 76.4 & 72.8 &
-0.7 & 91.0 & 85.6 & 83.8 & 87.4 & +6.1 & 29.5 & 26.3 & 49.8 & 32.3 & +5.4 \\
\coolname-UOIS$^\dagger$
& 75.2 & 62.4 & 68.8 & 68.8 & -16.0
& 74.8 & 54.1 & 68.0 & 65.6 & -7.9
& 81.1 & 71.0 & 72.7 & 74.9 & -6.4
& 65.3 & 56.9 & 55.9 & 59.4 & +32.5 \\
\rowcolor{lightgreenrow}
\coolname-UOIS$^\ddagger$
& 88.6 & 81.4 & 84.8 & 84.9 & +0.1
& 84.4 & 65.3 & 78.5 & 76.1 & +2.6
& 91.0 & 85.6 & 84.2 & 86.9 & +5.6
& 69.3 & 63.8 & 72.4 & 68.5 & +41.6 \\
\midrule
\multicolumn{21}{c}{\textit{Stage-2 (Frozen)}} \\
\midrule
MSMFormer~\cite{msmformer}
& 90.6 & 86.4 & 83.2 & 86.7 & 0
& 85.2 & 66.9 & 77.1 & 76.4 & 0
& 88.4 & 84.5 & 77.3 & 83.4 & 0
& 28.3 & 28.7 & 30.9 & 29.3 & 0 \\
Lu et al.~\cite{lu2023sss} (OCID) & 90.5 & 85.2 & 84.2 & 87.1 & +0.4 & 83.7 & 65.2 & 75.1 & 74.6 &
-1.8 & 89.7 & 84.9 & 79.1 & 85.6 & +2.2 & 24.7 & 28.3 & 28.4 & 26.9 & -2.4 \\
Lu et al.~\cite{lu2023sss} (OSD)  & 90.5 & 85.3 & 85.2 & 87.4 & +0.7 & 84.2 & 68.7 & 75.1 & 76.2 &
-0.2 & 91.7 & 87.8 & 86.0 & 89.0 & +5.6 & 35.7 & 39.3 & 49.1 & 39.9 & +10.6 \\
\coolname-UOIS$^\dagger$
& 83.2 & 77.4 & 68.8 & 76.5 & -10.2
& 80.0 & 61.1 & 68.0 & 69.7 & -6.7
& 86.9 & 82.2 & 72.7 & 80.6 & -2.8
& 62.7 & 56.4 & 55.9 & 58.3 & +29.0 \\
\rowcolor{lightgreenrow}
\coolname-UOIS$^\ddagger$
& 91.0 & 87.1 & 84.8 & 87.6 & +0.9
& 85.6 & 68.3 & 78.5 & 77.5 & +1.1
& 91.1 & 88.1 & 84.2 & 87.8 & +4.4
& 66.5 & 64.4 & 72.4 & 67.8 & +38.5 \\
\bottomrule
\end{tabular}}
\caption{
Longer training (50K iterations) on aggregated HumanPlay data.
The main paper uses iterative 2K-iteration fine-tuning.
{\coolname}-UOIS$^\dagger$: trained from scratch on HumanPlay data.
{\coolname}-UOIS$^\ddagger$: pretrained MSMFormer fine-tuned on HumanPlay data.
Stage-2 is frozen; its gains come solely from improved Stage-1.}
\label{tab:iteach_stage1_stage2}
\end{table*}
\endgroup

Training from scratch ({\coolname}-UOIS$^\dagger$) improves on the
HumanPlay test set ($\Delta C = +32.5$) but drops on OCID, OSD, and
RobotPushing, indicating that the HumanPlay data alone cannot replace
large-scale pretraining. Fine-tuning from pretrained weights
({\coolname}-UOIS$^\ddagger$) retains or improves performance on all four
benchmarks: $\Delta C$ is $+0.1$ on OCID, $+2.6$ on OSD, $+5.6$ on
RobotPushing, and $+41.6$ on HumanPlay for Stage-1.

Stage-2 is frozen throughout these experiments. The Stage-2 gains
($+0.9$, $+1.1$, $+4.4$, $+38.5$) are positive on all four benchmarks,
indicating that the Stage-1 improvements propagate through the pipeline.

  \subsection{System Configuration and Interface}
\label{app:interface}

\subsubsection{Experiment Setup}
\label{app:setup}

The system consists of a Fetch robot, an onboard laptop (RTX~4090,
16\,GB VRAM), and a HoloLens~2 headset (Fig.~\ref{fig:sys-components}).
The laptop handles inference, data logging, and fine-tuning; the robot
runs the ROS master; the HoloLens communicates via the ROS TCP
endpoint connector\footnote{\url{https://github.com/Unity-Technologies/ROS-TCP-Connector}}.
Table~\ref{tab:system_setup} lists the full configuration.

\begin{table}[H]
\centering
\renewcommand{\arraystretch}{1.15}
\resizebox{\linewidth}{!}{
\begin{tabular}{@{}ll@{}}
\toprule
\rowcolor{headerpeach}
\textbf{Component} & \textbf{Description} \\
\midrule
\rowcolor{gray!8}
\multicolumn{2}{@{}l}{\textit{Hardware}} \\
Robot & Fetch mobile manipulator with RGB-D sensing \\
Onboard compute & Lenovo Legion laptop with RTX 4090 GPU (16 GB VRAM) \\
MR interface & Microsoft HoloLens 2 for gaze and voice based annotation and visualization \\
\midrule
\rowcolor{gray!8}
\multicolumn{2}{@{}l}{\textit{Networking}} \\
Robot--Laptop link & Wired Ethernet for RGB-D data and perception outputs \\
HoloLens--Laptop link & Wi-Fi hotspot hosted on laptop for MR streaming \\
\midrule
\rowcolor{gray!8}
\multicolumn{2}{@{}l}{\textit{Software}} \\
Middleware & ROS with ROS-TCP connector for HoloLens communication \\
GPU usage & SAM2~\cite{ravi2024sam2} propagation, perception inference, and fine-tuning \\
Software stack & Ubuntu 22.04, Python 3.8, PyTorch 2.4, CUDA 12.1 \\
MR application & iTeachSkillsApp\footnotemark\ (Unity, UWP) for HoloLens 2 \\
UOIS toolkit & uois\_toolkit~\cite{uois_toolkit} (PyTorch) for training and evaluation \\
\bottomrule
\end{tabular}}
\caption{System configuration.}
\label{tab:system_setup}
\end{table}
\footnotetext{\url{https://github.com/IRVLUTD/iTeachSkillsApp}}

RGB-D frames from the robot are sent to the laptop for inference. Predictions
are streamed to the HoloLens, where the user provides gaze-based point
annotations on detected failures. Annotations are returned to the laptop for
SAM2~\cite{ravi2024sam2} mask propagation, after which the perception model
is fine-tuned onboard and redeployed. All components run on the robot's
onboard hardware; no external compute is required.

\subsubsection{Voice Commands}

Table~\ref{tab:voice_commands} lists the 10 voice commands grouped by
pipeline stage. All commands are recognized by the HoloLens speech engine;
no manual input is needed.

\begin{table}[H]
\centering
\renewcommand{\arraystretch}{1.2}
\resizebox{\linewidth}{!}{%
\begin{tabular}{@{}lllp{5.2cm}@{}}
\toprule
\rowcolor{headerpeach}
\textbf{Group} & \textbf{Pipeline Stage} & \textbf{Command} & \textbf{Action} \\
\midrule
Visualization
  & Failure observation
  & \texttt{Stream}       & Start live RGB-D streaming with predicted masks and bounding boxes overlaid on the HoloLens display \\
\midrule
\rowcolor{gray!8}
Capture
  & HumanPlay recording
  & \texttt{Begin Capture} & Start recording an RGB-D video sequence \\
\rowcolor{gray!8}
  && \texttt{Stop Capture}  & End the current recording \\
\midrule
Annotation
  & Final-frame labeling
  & \texttt{True Label}    & Register gaze point as a positive prompt \\
  && \texttt{False Label}   & Register gaze point as a negative prompt \\
  && \texttt{Erase Label}   & Remove all point prompts for the current object \\
  && \texttt{Next Object}   & Advance to the next object \\
\midrule
\rowcolor{gray!8}
Control
  & Mask propagation
  & \texttt{Send Label}    & Submit prompts to SAM2~\cite{ravi2024sam2} for bounding-box generation \\
\rowcolor{gray!8}
  & Data persistence
  & \texttt{Save Label}    & Persist annotated data to disk for training \\
\rowcolor{gray!8}
  & Session management
  & \texttt{Stop Label}    & End the annotation session \\
\bottomrule
\end{tabular}}
\caption{Voice commands in the {\coolname} MR interface, mapped to pipeline
stages. The user's hands remain free throughout.}
\label{tab:voice_commands}
\end{table}

\subsubsection{Practical Challenges}
\label{supp-sec: pract-exp-challenges}

Table~\ref{tab:practical_challenges} summarizes engineering challenges
encountered during deployment. These are system-integration issues and do
not affect the method itself. Fig.~\ref{fig:practical-challenges-encountered}
shows representative examples.

\begin{table}[H]
\centering
\renewcommand{\arraystretch}{1.2}
\rowcolors{2}{gray!8}{white}
\resizebox{\linewidth}{!}{
\begin{tabular}{@{}lll@{}}
\toprule
\rowcolor{headerpeach}
\textbf{Category} & \textbf{Challenge} & \textbf{Mitigation} \\
\midrule
Power & Laptop battery drain during GPU training & External power / periodic charging \\
Power & HoloLens battery limitation & External power bank during long sessions \\
Thermal & HoloLens overheating & Short pauses between sessions \\
Deployment & Unity / UWP build compatibility & Fixed dev environment or pre-built app \\
Networking & Wi-Fi hotspot + ROS communication & Onboard laptop as centralized hub \\
Compute & Concurrent inference + training & Onboard RTX 4090 GPU \\
\bottomrule
\end{tabular}}
\caption{Practical deployment challenges and mitigations.}
\label{tab:practical_challenges}
\end{table}

\begin{figure}[H]
    \centering
    \includegraphics[width=\linewidth]{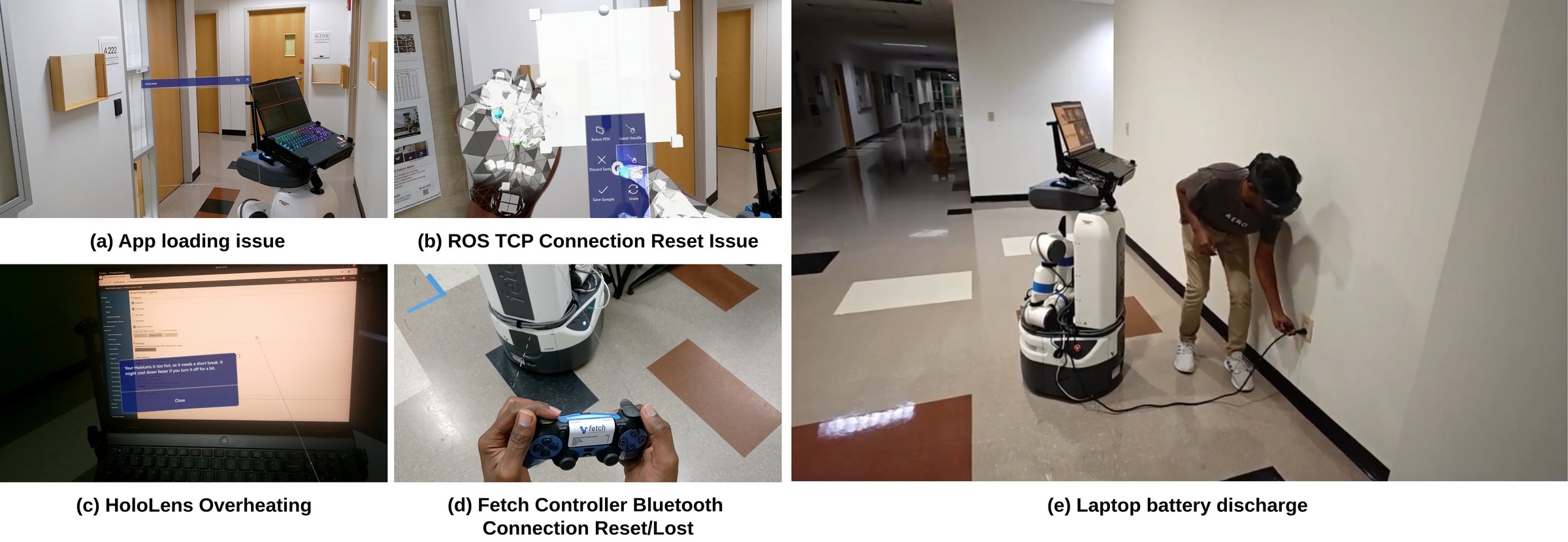}
    \vspace{4pt}
    \captionof{figure}{Practical challenges encountered during deployment.}
    \label{fig:practical-challenges-encountered}
\end{figure}

\begin{figure}[H]
    \centering
    \begin{subfigure}{0.48\linewidth}
        \includegraphics[width=\linewidth]{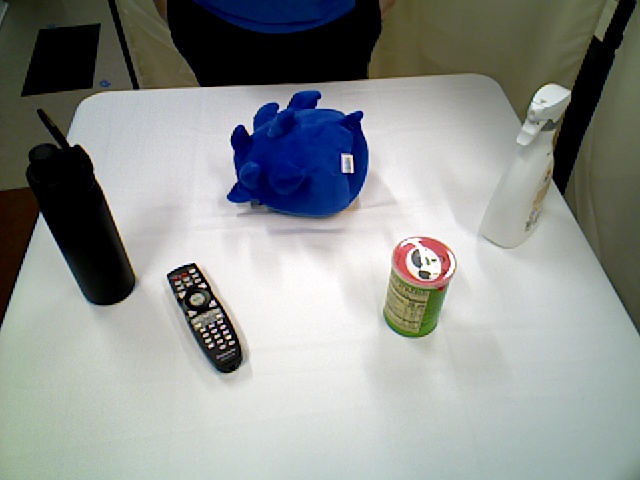}
        \caption*{\scriptsize White, scene 1}
    \end{subfigure}\hfill
    \begin{subfigure}{0.48\linewidth}
        \includegraphics[width=\linewidth]{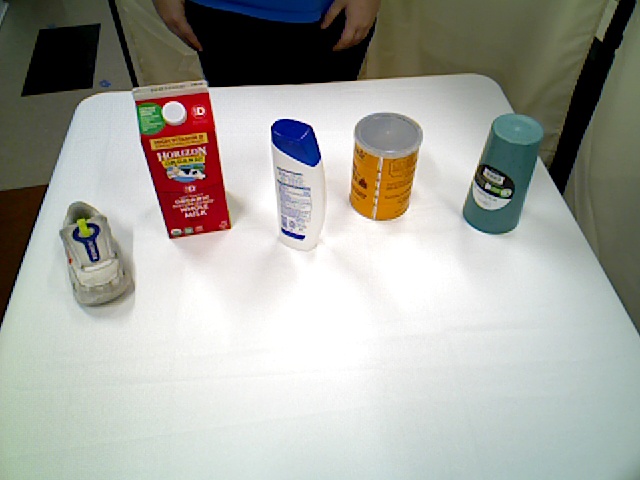}
        \caption*{\scriptsize White, scene 2}
    \end{subfigure}\\[2pt]
    \begin{subfigure}{0.48\linewidth}
        \includegraphics[width=\linewidth]{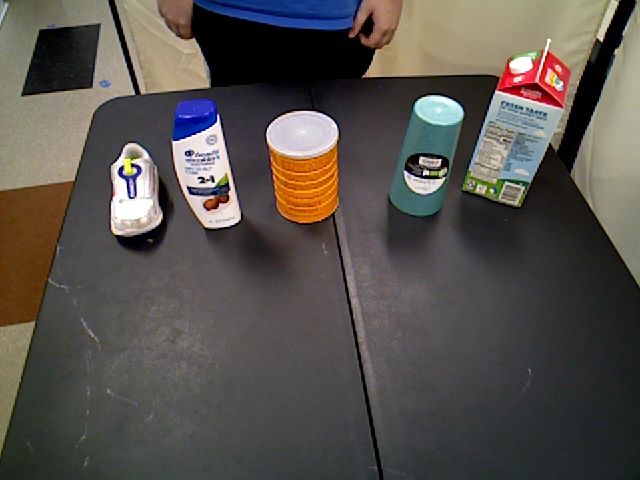}
        \caption*{\scriptsize Black, scene 1}
    \end{subfigure}\hfill
    \begin{subfigure}{0.48\linewidth}
        \includegraphics[width=\linewidth]{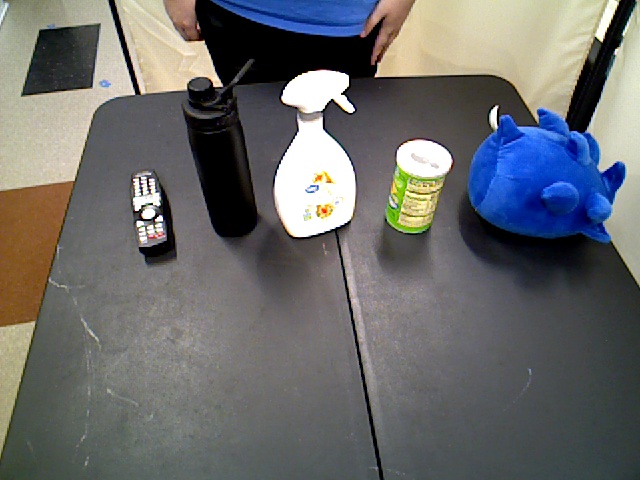}
        \caption*{\scriptsize Black, scene 2}
    \end{subfigure}
    \vspace{2pt}
    \caption{Four scenes annotated by P7, two per table surface.}
    \label{fig:study-setups}
\end{figure}

  \subsection{MR User Study Setups}
\label{app:user-study}

This appendix documents the setups and annotations from the 12-participant
MR user study (Sec.~\ref{subsec:user-study}). Each participant annotated
four scenes using two fixed sets of five household objects on two table
surfaces, white laminate and matte black, creating a contrast shift within
each session (Fig.~\ref{fig:study-setups}). The same object sets were used
across all participants; only the spatial layout was varied between
sessions. The intended presentation order alternated white-first and
black-first across consecutive participants; in practice, some sessions
deviated due to experimenter setup (see corner tags in
Fig.~\ref{fig:study-participants}).

Fig.~\ref{fig:study-participants} shows all 48 annotated frames.
Table~\ref{tab:study-summary} reports the per-group breakdown.
Expert and non-expert means are within one standard error on all
three measures.

\begin{table}[H]
\centering
\renewcommand{\arraystretch}{1.15}
\rowcolors{2}{gray!8}{white}
\resizebox{\linewidth}{!}{%
\begin{tabular}{@{}lccc@{}}
\toprule
\rowcolor{headerpeach}
\textbf{Measure} & \textbf{All ($n{=}12$)} & \textbf{Expert ($n{=}6$)} & \textbf{Non-expert ($n{=}6$)} \\
\midrule
Box IoU (\%) & 94.87 (0.49) & 94.95 (0.40) & 94.78 (0.59) \\
Completion time (s) & 351.19 (36.74) & 354.22 (36.86) & 348.16 (39.85) \\
NASA-TLX (/100) & 21.04 (12.56) & 20.42 (9.88) & 21.67 (15.77) \\
HumanPlay time (s) & \multicolumn{3}{c}{14.17 (3.15)} \\
Prompts / object & \multicolumn{3}{c}{1.03 (median 1, max 3)} \\
\bottomrule
\end{tabular}}
\caption{User study results. Values are mean (SD). HumanPlay time
and prompts per object are reported as aggregates across all participants.}
\label{tab:study-summary}
\end{table}

\begin{figure*}[p]
    \centering
    \includegraphics[width=\textwidth]{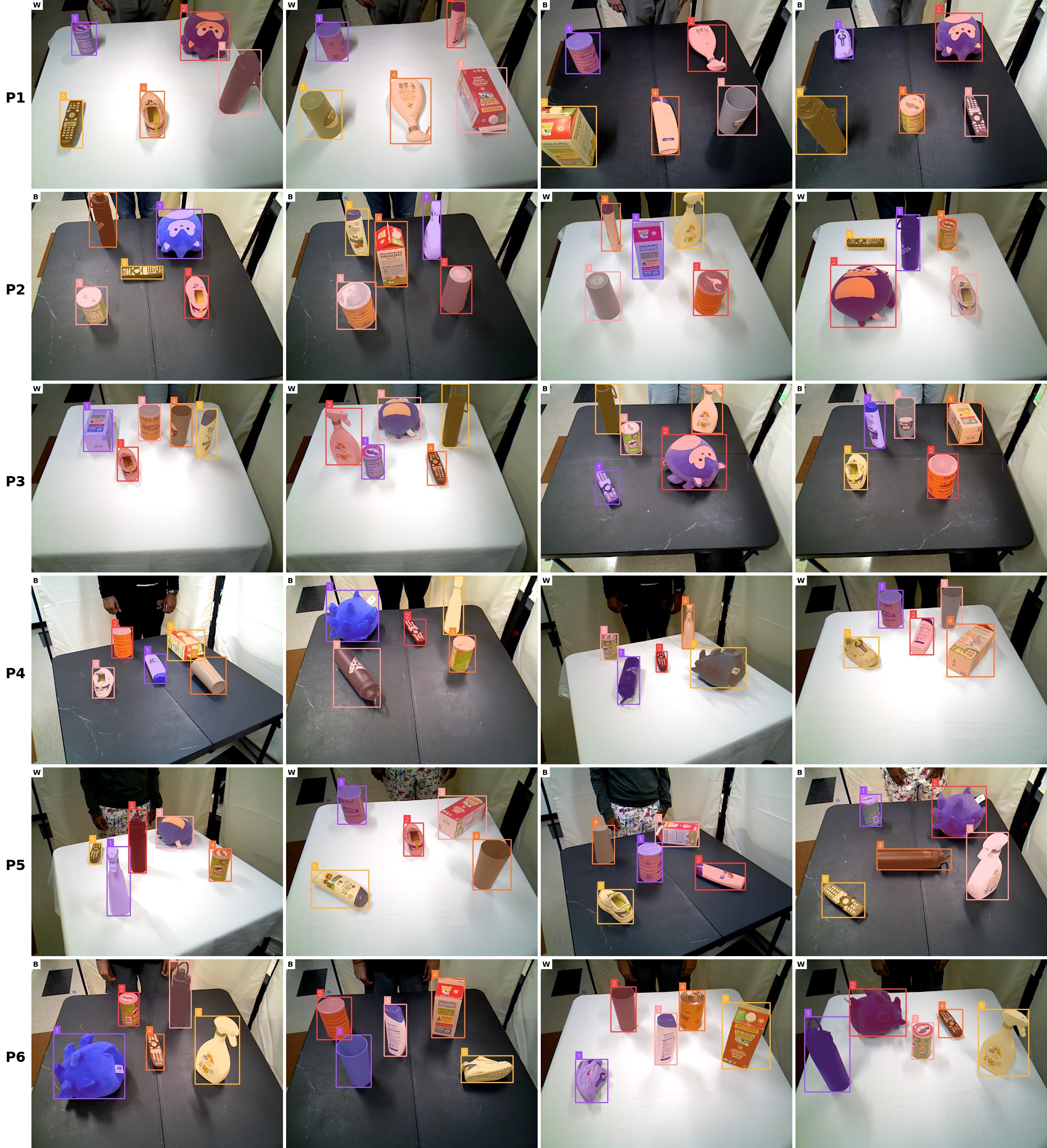}
    \caption{Per-participant annotations, P1--P6 (continued on next page).}
    \label{fig:study-participants-a}
\end{figure*}

\begin{figure*}[p]
    \centering
    \includegraphics[width=\textwidth]{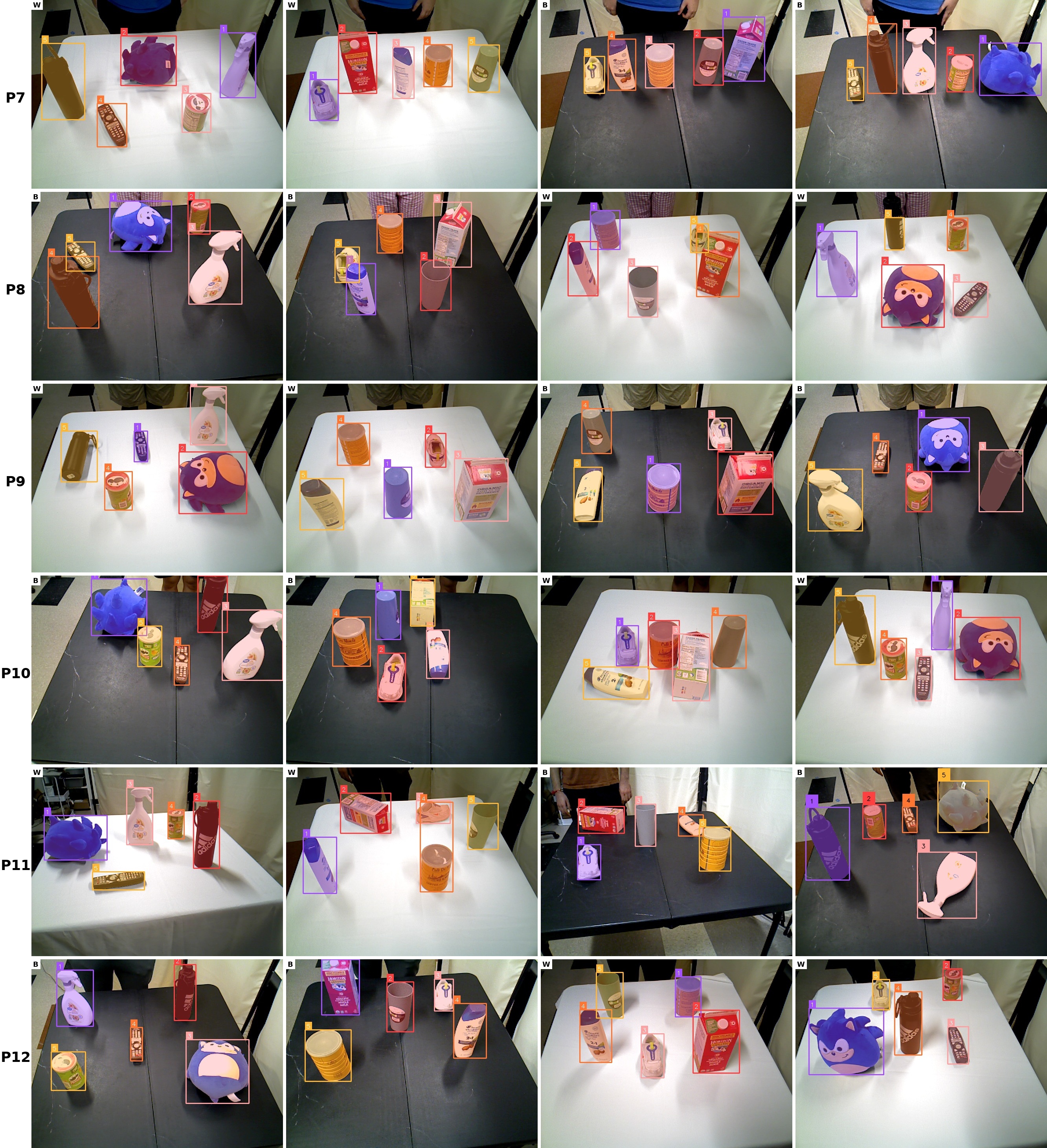}
    \vspace{6pt}
    \caption{Per-participant annotations, P7--P12 (continued from previous
    page). All 48 annotated frames with bounding boxes and instance masks
    numbered in labeling order.
    Odd participants (P1, P3, P5, P7, P9, P11) follow W--W--B--B order;
    even participants (P2, P4, P6, P8, P10, P12) follow B--B--W--W.
    Corner tags indicate table color.}
    \label{fig:study-participants}
\end{figure*}

\fi

\end{document}